\pdfoutput=1
\documentclass[11pt]{article}

\usepackage[preprint]{acl}

\usepackage{times}
\usepackage{latexsym}
\usepackage{tabularx}
\usepackage[T1]{fontenc}

\usepackage[utf8]{inputenc}

\usepackage{microtype}

\usepackage{inconsolata}

\usepackage{graphicx}
\usepackage{booktabs}
\usepackage{amsmath}
\usepackage{amssymb}
\usepackage{amsfonts}
\usepackage{enumitem}
\usepackage{hyperref}
\usepackage{cleveref}

\usepackage[table]{xcolor}
\definecolor{epiworldblue}{RGB}{220,235,250}
\definecolor{epiworldblueedge}{RGB}{130,170,210}

\usepackage[most]{tcolorbox}
\tcbuselibrary{listings}
\definecolor{epiworldcode}{RGB}{236,240,244}
\definecolor{epiworldcodeedge}{RGB}{160,170,185}

\newtcolorbox{promptbox}[1][]{%
  enhanced,
  colback=epiworldblue,
  colframe=epiworldblueedge,
  arc=4pt,
  boxrule=0.5pt,
  left=8pt,
  right=8pt,
  top=6pt,
  bottom=6pt,
  title={\small #1},
  fonttitle=\bfseries,
  attach boxed title to top left={xshift=8pt, yshift=-7pt},
  boxed title style={
    colback=epiworldblueedge,
    colframe=epiworldblueedge,
    sharp corners,
    arc=2pt,
    boxrule=0pt
  }
}

\newtcblisting{codebox}[1][]{%
  enhanced,
  listing only,
  colback=epiworldcode,
  colframe=epiworldcodeedge,
  arc=4pt,
  boxrule=0.4pt,
  left=8pt,
  right=8pt,
  top=4pt,
  bottom=4pt,
  title={\small #1},
  fonttitle=\bfseries,
  listing options={
    basicstyle=\ttfamily\footnotesize,
    breaklines=true,
    columns=fullflexible,
    keepspaces=true,
    showstringspaces=false
  },
  attach boxed title to top left={xshift=8pt, yshift=-7pt},
  boxed title style={
    colback=epiworldcodeedge,
    colframe=epiworldcodeedge,
    sharp corners,
    arc=2pt,
    boxrule=0pt
  }
}

\newcommand{\promptag}[1]{\colorbox{epiworldblueedge!40!white}{\texttt{\small #1}}}

\graphicspath{{figures/}}

\newcommand{\eg}{e.g.,\ }

\title{EpiWorld: Grounding LLM Policy Agents in Epidemiological World Models}

\author{
  \textbf{Zeeshan Memon\textsuperscript{1}},
  \textbf{Yiqi Su\textsuperscript{2}},
  \textbf{Kai Shu\textsuperscript{1}},
  \textbf{Naren Ramakrishnan\textsuperscript{2}},
  \textbf{Liang Zhao\textsuperscript{1}}
\\
\\
  \textsuperscript{1}Department of Computer Science, Emory University, Atlanta, GA, USA
\\
  \textsuperscript{2}Department of Computer Science, Virginia Tech, Alexandria, VA, USA
}

\begin{document}
\maketitle

\begin{abstract}
Epidemic intervention policies are textual artefacts that human decision-makers interpret, justify, and revise through natural language, making large language models a natural candidate for epidemic policy reasoning. A naive LLM, however, lacks the epidemic dynamics needed to project intervention consequences, the quantitative surveillance signals required to assess severity, and the institutional constraints that define admissible actions. We present EpiWorld, a closed-loop framework that grounds an LLM policy actor in a learned action-conditioned epidemiological world model and a tiered skill library of public-health protocols, surveillance tools, and adaptive lessons accumulated through after-action analysis. Given a candidate intervention, the world model predicts regional epidemic evolution and enables fast counterfactual rollouts that provide feedback for policy selection and refinement. Outcomes of simulated futures are distilled into reusable lessons while protocol constraints remain fixed, allowing the decision process to improve without sacrificing interpretability or controllability. We evaluate both the world model and the end-to-end framework on retrospective COVID-19 and Influenza datasets: the world model achieves the best out-of-distribution Peak-MAE among all forecasting baselines, and the closed-loop framework reduces cumulative hospitalisation by up to 59\% across datasets and by an average of $\sim$16\% across six LLM backbones, outperforming reinforcement-learning and optimal-control policy baselines.
\end{abstract}

\section{Introduction}
\label{sec:intro}
\begin{figure*}
    \centering
\includegraphics[width=0.95\linewidth, trim={0 15cm 0 0}, clip]{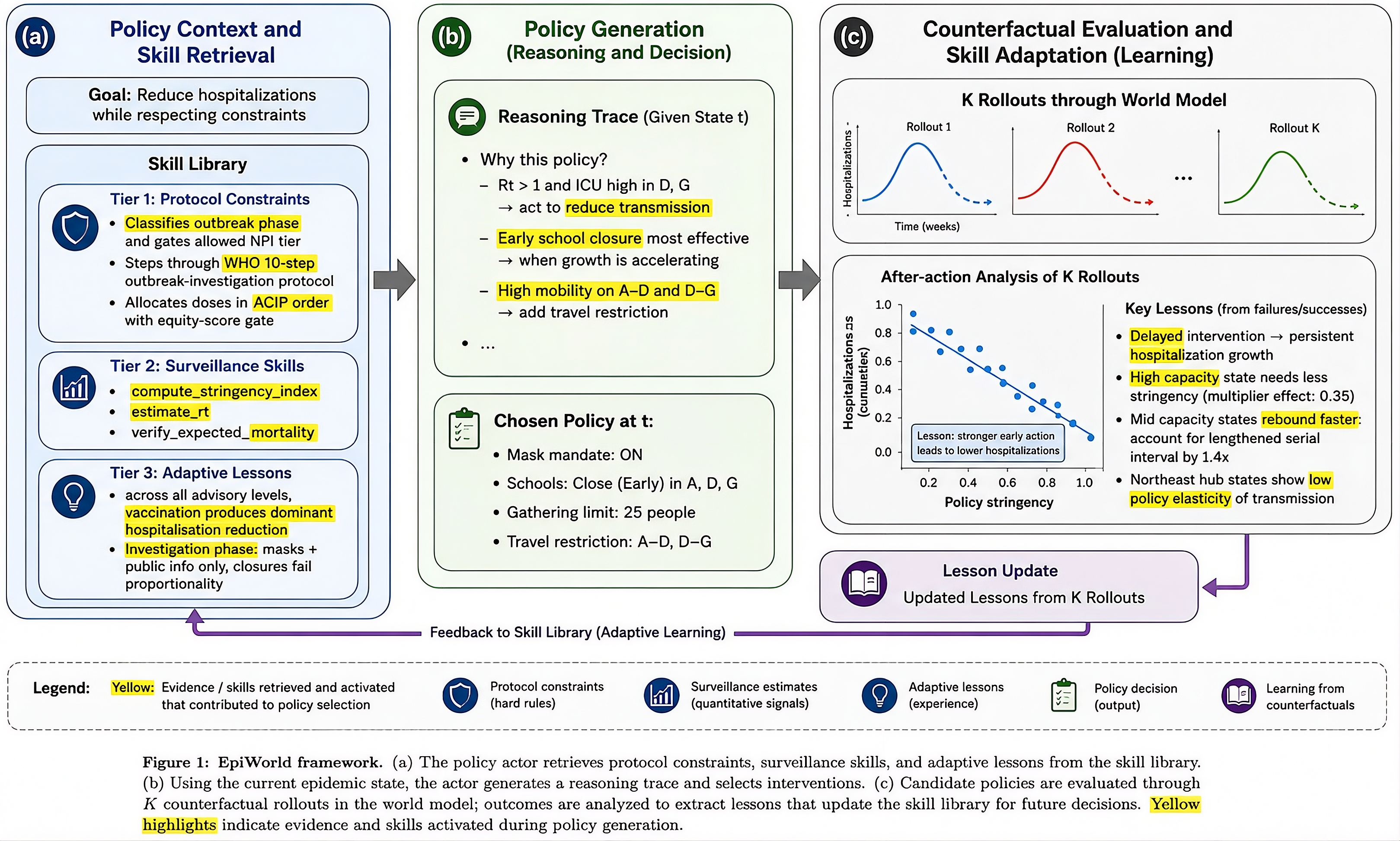}
    \caption{EpiWorld framework. (a) The policy actor retrieves protocol constraints, surveillance skills, and adaptive lessons. (b) The actor generates interventions and reasoning traces. (c) Counterfactual rollouts evaluate candidate policies and produce updated lessons. Yellow highlights denote retrieved evidence and activated skills.}
    \label{fig:method_fig}
\end{figure*}
Many policy reasoning problems are language-mediated: the relevant
knowledge, constraints, and decisions are expressed through
natural-language artefacts such as statutes, regulations, advisories,
and operational guidance. Epidemic response is one such setting.
Public-health authorities communicate interventions through textual
policies while simultaneously integrating surveillance observations,
forecasting signals, and previously enacted decisions under
uncertainty. Because policies influence future observations and must
themselves be justified and communicated through language, epidemic
response forms a natural testbed for studying whether large language
models can act as policy reasoners when grounded in quantitative
decision environments.

Epidemic response is fundamentally a problem of decision making over
counterfactual futures. Public-health authorities must not only forecast
outcomes under current policies, but also reason about what would happen
under alternative interventions and which actions should be taken next
to optimise public-health objectives such as hospital burden and
mortality~\citep{flaxman2020nature,hale2021}. During an evolving outbreak,
these decisions are made repeatedly across interacting regions under
delayed clinical observations~\citep{gibson2020mechbayes}, shifting behavioural
responses~\citep{williams2023generativeagents,park2023generativeagents}, and policy-endogenous
dynamics, where observed trajectories are themselves consequences of
prior interventions~\citep{memon2026worldmodels}. As a result, epidemic
policy optimisation requires models that can evaluate alternative
intervention trajectories rather than forecast a single future.

Existing computational tools only partially address this challenge.
Statistical and neural epidemic forecasters can predict future
hospitalisations, or deaths under observed
covariates~\citep{shi2024pancode,wan2025earth,du2024pandemicllm},
but they are not designed to answer counterfactual policy questions:
they estimate what is likely to happen, not what would happen under
alternative interventions. Causal-inference methods estimate
retrospective intervention effects~\citep{flaxman2020nature}, yet struggle
with co-occurring policies, delayed outcomes, and prospective
multi-step decision making. Mechanistic simulators and
reinforcement-learning approaches support policy optimisation in
principle~\citep{williams2023generativeagents,chopra2025agency,bampa2022epidrlearn,song2020durleca,epipolicyrl2023},
but depend on predefined dynamics or simulation environments that are
difficult to adapt, inspect, and revise. Conversely, large language
models operate over textual protocols, advisories, and accumulated
lessons, yet lack grounded epidemic dynamics for evaluating the
downstream consequences of their recommendations.

{\bf The missing object is therefore a grounded language-mediated decision environment: one that allows an LLM policy actor to reason over textual guidance while testing proposed actions against learned policy-conditioned epidemic dynamics.} In machine learning, this object is known as a \emph{world model}~\citep{hafner2021dreamerv2,hafner2023dreamerv3,micheli2023iris,hansen2022tdmpc}: a learned dynamics model that supports planning and policy learning through imagined rollouts~\citep{hafner2019planet,hafner2020dreamer}. World models have shown strong performance in games and robotics, where agents improve decisions by simulating future consequences before acting. Epidemic world modelling~\citep{memon2026worldmodels}, however, is not a direct transplant of these settings: epidemic dynamics evolve across interacting regions, delayed clinical signals, and intervention-dependent behavioural responses that alter the observed trajectory itself~\citep{flaxman2020nature,kang2020multiscale}. Moreover, epidemic interventions are textual and institutional objects rather than unconstrained actions---they must be justified, communicated, and constrained under public-health protocols~\citep{hale2021}.

We introduce EpiWorld, a closed-loop framework for epidemiological policy optimisation with LLM agents (Figure~\ref{fig:method_fig}). EpiWorld couples an action-conditioned regional world model with a protocol-guided policy actor. As illustrated in Figure~\ref{fig:method_fig}(a), the actor first constructs policy context through a tiered epidemiological skill library, consisting of protocol constraints, surveillance tools, and adaptive lessons accumulated from prior experience, mirroring the components of the standard public-health monitoring and evaluation cycle~\citep{who2018ihrmef}. Conditioned on this context, the actor generates interventions and explicit reasoning traces over the current epidemic state (Figure~\ref{fig:method_fig}(b)). Candidate policies are then evaluated through the regional world model, which predicts how epidemic trajectories evolve under interventions and enables fast counterfactual rollout across interacting regions (Figure~\ref{fig:method_fig}(c)). Outcomes of simulated futures are retrospectively analysed to update the lesson library and improve future policy generation. Thus, the LLM is not an unconstrained recommender but a language-facing policy reasoner whose proposals remain protocol-constrained and grounded through counterfactual epidemic rollouts, and can assist public-health authorities as a decision-support layer whose recommendations remain subject to their expert approval.
Our contributions are as follows:
\begin{itemize}[leftmargin=*,itemsep=0pt,topsep=0pt,parsep=0pt]
\item We formulate epidemic policy optimisation as grounded, language-mediated, closed-loop decision making unifying forecasting, counterfactual reasoning, and prospective intervention optimisation within a single rollout-based framework.
\item We propose an epidemiological world model for regional epidemic dynamics that integrates delay-aware temporal encoding, graph-structured regional coupling, residual next-state prediction, and policy-conditioned latent modulation.
\item We introduce a protocol-guided LLM policy actor that operationalises public-health protocols, surveillance estimators, and after-action lessons as an interpretable and human-intervenable epidemic decision mechanism.
\end{itemize}
Together, these results suggest a new direction for epidemic AI systems: not merely predicting outbreaks, and not merely prompting LLMs to recommend interventions, but using LLMs as the connective layer between textual policy reasoning and
epidemiologically grounded imagined futures.

\section{Related Work}
\label{sec:related}

\paragraph{Forecasters and LLM proposers.}
Mechanistic Bayesian models dominate classical epidemic forecasting
\citep{flaxman2020nature,gibson2020mechbayes,qian2020compartmentalgp},
and neural extensions replace hand-coded compartments with latent ODEs
and graph-CDE encoders~\citep{shi2024pancode,wan2025earth}. LLMs have
since entered the loop as trend forecasters
\citep{du2024pandemicllm,yu2026rwml}, policy proposers inside
mechanistic environments
\citep{aoki2026policymakers,mao2025epiplanagent,coordinated2026pandemic},
agentic simulator builders~\citep{datta2026epiagent}, grounded
narrative generators over scenario
projections~\citep{datta2026epinarrate}, which communicate projections
rather than optimise interventions, and generative
behavioural agents~\citep{williams2023generativeagents,vacsim2025,
park2023generativeagents}. These methods either fold interventions
into covariate weightings or evaluate policies inside a fixed
mechanistic simulator. We retain the LLM as the policy proposer and
close the loop through a learned world model.

\paragraph{Policy optimisation and skill learning.}
Optimal control over compartmental or metapopulation dynamics yields
analytic policies under assumed dynamics
\citep{alvarez2021simple,acemoglu2021multigroup,bertsimas2022delphi},
and deep RL trains policies on simulated epidemic environments
\citep{bampa2022epidrlearn,song2020durleca} but transfers poorly across
disease or period. LLM-agent work reframes policy improvement as skill
accumulation~\citep{wang2023voyager,shinn2023reflexion,xia2026skillrl}
and prompt optimisation
\citep{agrawal2025gepa,khattab2023dspy,yang2023opro}, with LLM-driven
simulators closing the loop~\citep{chopra2025agency,xi2025agentgymrl}.
We instantiate this direction for epidemic intervention, accumulating
epidemic-specific skills against a learned world model.

\paragraph{World models.}
World models learn latent dynamics that support policy learning through
in-model rollouts~\citep{hafner2019planet,hafner2020dreamer}, refined
via discrete representations, transformer backbones, and
model-predictive integration
\citep{hafner2021dreamerv2,hafner2023dreamerv3,micheli2023iris,
hansen2022tdmpc}. This literature is almost exclusively developed in
robotics and games, with recent extensions to structured scientific
domains such as actionable molecular
optimisation~\citep{qiao2026molworld}. \citet{memon2026worldmodels} formalise the
epidemic case as a controlled, partially-observed system with
policy-endogenous observations. We instantiate that framing as an
action-conditioned recurrent state-space world model and use it as the
rollout substrate for our policy.

\section{Method}
\label{sec:method}
\subsection{Problem Formulation}
\label{sec:problem}

We consider epidemic policy optimisation over a set of interacting regions.
Let $\mathcal{G}=(\mathcal{V},\mathcal{E})$ denote a regional interaction
graph, where each node $i\in\mathcal{V}$ represents a geographic unit and
each edge $(i,j)\in\mathcal{E}$ encodes epidemiologically relevant coupling,
such as mobility, spatial adjacency, commuting flow, or correlated
transmission pressure.
Each region is associated with static attributes
$\mathbf{s}_i$ describing demographic, socioeconomic, healthcare-capacity,
and vulnerability indicators, and dynamic epidemic observations
$\mathbf{x}_i^t$, including infections, hospitalisations, mortality,
vaccination uptake, testing intensity, and disease severity signals.
Aggregating across regions gives the global state
$\mathbf{X}^t=\{\mathbf{x}_i^t\}_{i\in\mathcal{V}}$
and static context
$\mathbf{S}=\{\mathbf{s}_i\}_{i\in\mathcal{V}}$.

At each decision step $t$, a policy maker selects interventions
$\mathbf{a}^t=\{\mathbf{a}_i^t\}_{i\in\mathcal{V}}$ with
$\mathbf{a}_i^t\in\mathcal{A}$, where $\mathcal{A}$ may include
non-pharmaceutical interventions, vaccination allocation,
school policies, testing strategies, mobility restrictions,
or other public-health actions. The epidemic evolves according to unknown policy-conditioned dynamics:
\begin{equation}
\mathbf{X}^{t+1}
\sim
P\!\left(
\cdot
\mid
\mathbf{X}^{t},
\mathbf{S},
\mathbf{a}^{t},
\mathcal{G}
\right).
\label{eq:true_dynamics}
\end{equation}
Given a policy objective $g$, such as minimising cumulative
hospitalisations, reducing peak healthcare burden, satisfying
intervention budgets, or enforcing equity constraints, the goal is to
find a goal-conditioned decision policy
$\pi(\mathbf{a}^{t}\mid\mathbf{X}^{t},\mathbf{S},\mathcal{G},g)$
that maximises expected utility over horizon $H$:

\begin{equation}
\pi^\star
=
\arg\max_{\pi}
\;
\mathbb{E}_{\tau\sim(\pi,P)}
\left[
J_g(\tau)
\right],
\label{eq:policy_objective}
\end{equation}
where
$\tau=(\mathbf{X}^{0},\mathbf{a}^{0},\mathbf{X}^{1},\mathbf{a}^{1},\ldots,\mathbf{X}^{H})$
denotes the resulting trajectory. This formulation is intentionally method-agnostic and encompasses
existing epidemic decision-making paradigms. Human public-health
operators incorporate contextual knowledge and operational constraints,
but decisions may be subjective and difficult to scale consistently.
Mechanistic simulators and optimal-control approaches support rich
counterfactual analysis but often depend on hand-designed assumptions
and become computationally expensive for large-scale policy search.
Reinforcement-learning approaches can optimise long-horizon
interventions but typically require simulator access and may offer
limited semantic feedback or controllability for domain experts.
Surrogate forecasting models are computationally efficient but are
primarily optimised for prediction rather than iterative policy
evaluation. These limitations motivate the need for models that can combine
expressive epidemic dynamics, efficient counterfactual evaluation,
and scalable policy exploration.

\subsection{EpiWorld Framework}
\label{sec:framework}

EpiWorld addresses the epidemic policy optimisation problem in
Eq.~\ref{eq:policy_objective} through two coupled components:
(i) an epidemiological world model that approximates the unknown
policy-conditioned dynamics and enables counterfactual rollout, and
(ii) a protocol-guided policy actor that generates interventions from
predicted epidemic trajectories. Formally, EpiWorld introduces a world model that learns unknown epidemic dynamics
\begin{equation}
f_\theta
:
(\mathbf X^t,\mathbf S,\mathbf a^t,\mathcal G)
\rightarrow
\hat{\mathbf X}^{t+1},
\qquad
f_\theta \approx P,
\label{eq:framework_world}
\end{equation}
which predicts policy-conditioned epidemic evolution and supports
multi-step counterfactual rollout. Policy optimisation is then performed through interaction with the
world model:
\begin{equation}
\pi^\star
\in
\arg\max_{\pi}
\;
\mathbb E_{
\hat{\tau}
\sim
(\pi,f_\theta)
}
\left[
J_g(\hat{\tau})
\right],
\label{eq:epiworld_obj}
\end{equation}
where
$\hat{\tau}
=
(
\hat{\mathbf X}^{1},
\hat{\mathbf a}^{1},
\ldots,
\hat{\mathbf X}^{H}
)$
denotes a simulated trajectory. The closed loop replaces the unknown
$P$ in Eq.~\ref{eq:policy_objective} with the learned $f_\theta$,
enabling forecasting, counterfactual evaluation, and policy
optimisation within a unified framework.

\subsection{Graph-Structured Epidemiological World Model}
\label{sec:wm}

Generic world models are typically designed for homogeneous sequential
prediction and do not explicitly account for delayed disease
progression, intervention feedback, or structured transmission across
interacting regions. To address these limitations, we specialise the
world-model paradigm for epidemic policy reasoning through three
epidemiological inductive biases: delay-aware temporal encoding,
policy-conditioned latent modulation, and graph-structured regional
coupling. The world model introduced in Eq.~\ref{eq:framework_world} adopts an
encoder--dynamics architecture trained through multi-objective
supervision in both latent and observation spaces. Epidemic observables evolve asynchronously: infections precede
reported cases, cases precede hospitalisations, and hospitalisations
precede mortality signals. To capture both short-term recency and
delayed cross-channel effects, we encode each regional trajectory
through two complementary temporal branches. For node $i\in\mathcal{V}$ with
$\mathbf{x}_i=\mathbf{X}_i^{\,t-W:t}$, where $W$ denotes the temporal
observation window,
\begin{equation}
\mathbf{e}^{\mathrm r}_i
=
\mathrm{GRU}_{\theta}(\mathbf{x}_i),
\;
\mathbf{e}^{\mathrm c}_i
=
\mathrm{CDE}_{\theta}
\!\left(
\mathrm{DelayMix}_{\theta}
(\mathbf{x}_i)
\right),
\label{eq:wm-temporal}
\end{equation}
where $\mathrm{DelayMix}$ is a causal cross-channel mixing module and
$\mathrm{CDE}$ models continuous temporal evolution along the induced
path. A learnable scalar gate $\gamma$ balances the contribution of the two
views, $\mathbf{e}_i = \left[\mathbf{e}^{\mathrm r}_i,\;
\sigma(\gamma)\mathbf{e}^{\mathrm c}_i\right]$, with $\sigma$ the
logistic function.
Actions modify how epidemic trajectories evolve rather than
acting as ordinary covariates. The temporal representation is
combined with static regional context and projected into a
policy-conditioned latent representation $\mathbf{z}_i^t$, which
serves as the internal state for epidemic dynamics modelling. We
condition regional latent states through feature-wise modulation,
\begin{equation}
\begin{aligned}
\mathbf{z}^{t}_i
&=
\left(
\mathbf{1}
+
\boldsymbol{\beta}^{t}_i
\right)
\odot
\mathrm{Proj}_{\theta}
\!\left(
[
\phi_s(\mathbf{s}_i),
\mathbf{e}_i
]
\right)
+
\boldsymbol{\alpha}^{t}_i,
\\
(
\boldsymbol{\alpha}^{t}_i,
\boldsymbol{\beta}^{t}_i
)
&=
\psi_\theta
(
\mathbf{a}^{t}_i
).
\end{aligned}
\label{eq:wm-encoder}
\end{equation}
Here $\mathbf{z}^{t}_i, \boldsymbol{\alpha}^{t}_i,
\boldsymbol{\beta}^{t}_i \in \mathbb{R}^{d_z}$, and
$\psi_\theta:\mathcal{A}\!\to\!\mathbb{R}^{2 d_z}$ is a small MLP that
maps the per-region action to FiLM-style scale and shift parameters.
This conditioning allows interventions to reshape every latent channel
rather than entering as additive inputs. Stacking over regions yields
$\mathbf z^t=(\mathbf z_i^t)_{i\in\mathcal V}$, which initialises
$\mathbf Z^{(0)}$. Epidemic evolution exhibits strong inter-region dependency through
mobility, adjacency, and correlated behavioural patterns. To model
regional transmission while preventing information leakage across
unrelated locations, we apply adjacency-masked graph attention:
\begin{equation}
\begin{aligned}
\mathbf Z^{(\ell+1)}
&=
\mathrm{FFN}
\!\left(
\mathrm{Attn}
(
\mathbf Z^{(\ell)};
\mathcal G
)
\right),
\\
[\mathrm{Attn}(\mathbf Z;\mathcal G)]_{ij}
&\propto
\mathbb 1[(i,j)\in\mathcal E]
\exp
\left(
\frac{
\mathbf q_i^\top
\mathbf k_j
}{
\sqrt d
}
\right).
\end{aligned}
\label{eq:wm-dynamics}
\end{equation}
After $L$ layers, the model predicts residual updates,
\begin{equation}
\hat{\mathbf z}^{\,t+1}
=
\mathbf Z^{(L)},
\qquad
\hat{\mathbf X}^{\,t+1}
=
\mathbf X^t
+
\mathrm{Head}_{\theta}
(
\hat{\mathbf z}^{\,t+1}
),
\label{eq:wm-decode}
\end{equation}
which stabilises learning for evolving epidemic trajectories
while preserving responsiveness to transmission shocks and
interventions. The world model is trained through complementary supervision
objectives. A latent transition objective aligns predicted latent
dynamics with encoded future representations,
\begin{equation}
\resizebox{\columnwidth}{!}{$
\mathcal L_{\mathrm{lat}}
=
\mathbb E_{\mathcal D_{\mathrm{hist}}}
\left[
\left\|
\hat{\mathbf z}^{\,t+1}
-
\mathrm{Enc}_{\theta}
(\mathbf X^{\,t+1-W:t+1}, \mathbf S, \mathbf a^{\,t})
\right\|^2
\right]$},
\label{eq:wm-lat}
\end{equation}
Here $\mathrm{Enc}_{\theta}$ denotes the same encoder used to
construct the latent state $\mathbf{z}^t$ in
Eq.~\ref{eq:wm-encoder}, applied to the next-step window to form the
prediction target; inference uses the predictor alone.
An observation objective additionally anchors predictions to measured
epidemic outcomes,
\begin{equation}
\mathcal L_{\mathrm{obs}}
=
\mathbb E_{\mathcal D_{\mathrm{hist}}}
\left[
\mathcal L_{\mathrm{rec}}
(
\hat{\mathbf X}^{\,t+1},
\mathbf X^{\,t+1}
)
\right],
\label{eq:wm-obs}
\end{equation}
where $\mathcal L_{\mathrm{rec}}$ is a per-channel Huber loss for
robustness to reporting noise and count outliers. The final objective
is $\mathcal L = \mathcal L_{\mathrm{lat}} + \lambda \mathcal L_{\mathrm{obs}}$.
Together, delay-aware temporal modelling, policy-conditioned latent
modulation, and graph-constrained regional interaction transform
generic world-model reasoning into an epidemic-specialised dynamics
model for counterfactual policy optimisation.

\subsection{Protocol-Guided LLM Policy Actor}
\label{sec:policy}

Predicted trajectories alone are insufficient for policy
making: decisions must remain protocol-compliant, draw on quantitative
surveillance, and adapt to behavioural feedback from prior responses.
EpiWorld therefore implements the policy $\pi(\cdot\mid g)$ as a
protocol-guided LLM actor equipped with an explicit skill library.
The actor produces interventions through a tiered skill hierarchy,
\begin{equation}
\pi
(
\mathbf a^t
\mid
\mathbf X^t,
\mathbf S,
\mathcal G,
g,
\mathcal T
),
\qquad
\mathcal T
=
\mathcal T_1
\cup
\mathcal T_2
\cup
\mathcal T_3,
\label{eq:policy}
\end{equation}
where the three tiers correspond to Protocol Constraints,
Surveillance Skills, and Adaptive Lessons. Each tier reflects a
distinct mode of public-health knowledge, regulatory protocols,
quantitative surveillance, and operational learning, grounded
respectively in the International Health Regulations~\citep{who2005ihr},
the Joint External Evaluation tool~\citep{who2018jee}, and WHO
After-Action Review guidance~\citep{who2019afterreview}; our
contribution exposes each layer as a skillset for agents.

\paragraph{Protocol Constraints ($\mathcal{T}_1$).}

Protocol Constraints are deterministic procedures codified by
international and national public-health authorities: pandemic-interval
classification, ACIP vaccine-allocation ordering, NPI proportionality
validation, and IHR notification thresholds. Each skill $\kappa\in\mathcal T_1$ returns a constraint
predicate $c_\kappa(\mathbf a, \mathbf X^t, g) \in \{0,1\}$.
The resulting admissible action envelope is
\begin{equation}
\mathcal A_g^t
=
\left\{
\mathbf a\in\mathcal A
\;\middle|\;
c_\kappa(
\mathbf a,
\mathbf X^t,
g
)=1,
\;
\forall
\kappa\in\mathcal T_1
\right\},
\label{eq:t1}
\end{equation}
which is computed once per decision step and remains fixed during
policy generation.

\paragraph{Surveillance Skills ($\mathcal{T}_2$).}

Surveillance Skills are analytical tools epidemiologists
use: $R_t$ estimators~\citep{cori2013rt} quantify transmission
acceleration, severity ratios (CFR, IFR) project hospital load, the
OxCGRT stringency index~\citep{hale2021} aggregates NPI levels into a
scalar, and the ATSDR Social Vulnerability Index flags vulnerable
populations. Each skill
$\kappa\in\mathcal T_2$
contains parameters
$\phi_\kappa\in B_\kappa$
and returns quantitative summaries
\begin{equation}
\hat y_\kappa
=
\kappa
(
\mathbf X^t,
\mathbf S,
\mathcal G;
\phi_\kappa
),
\label{eq:t2}
\end{equation}
which provide interpretable evidence for policy generation under the
admissible action set $\mathcal A_g^t$.
\paragraph{Adaptive Lessons ($\mathcal{T}_3$).}
Adaptive Lessons store reusable lessons accumulated from previous
world-model rollouts. This mirrors the After-Action Review framework
used by WHO and FEMA to convert past outbreaks into reusable
operational knowledge~\citep{who2019afterreview}. Each lesson is represented as
$\ell = (\tau_\ell, \rho_\ell, \boldsymbol\delta_\ell)$,
where $\tau_\ell$ denotes behavioural and regional tags,
$\rho_\ell\in[0,1]$ is a confidence score, and
$\boldsymbol\delta_\ell$ is a parameter adjustment targeting
Surveillance Skill parameters. At decision step $t$, the actor extracts tags from the current
context,
$\tau(\mathbf X^t,\mathbf S)$,
retrieves the top-$k$ matching lessons
$L^t\subset\mathcal T_3$,
and refines surveillance parameters through,
\begin{equation}
\phi'_\kappa
=
\mathrm{clip}_{B_\kappa}
\left(
\phi_\kappa
+
\sum_{\ell\in L^t}
\rho_\ell
\boldsymbol\delta_\ell^{(\kappa)}
\right),
\label{eq:t3-update}
\end{equation}
allowing lessons accumulated in one region to transfer to another
with a similar behavioural signature within the same infection
regime. Lessons are calibrated to the regime they are distilled from
and are not transferable across arbitrary regimes or world models
(\S\ref{sec:res:policy}). Adaptive
Lessons can refine Surveillance Skill estimates but cannot expand the
Protocol-Constraint admissible action envelope. At inference, the
actor applies $\mathcal{T}_1$ to fix $\mathcal{A}^t_g$, retrieves and
applies $\mathcal{T}_3$ lessons to refine $\mathcal{T}_2$ parameters,
and conditions on the resulting $\mathcal{T}_2$ estimates to propose
an action within $\mathcal{A}^t_g$. The examples of these skills are shown in Appendix~\ref{app:skills}; constraint-satisfaction rates over real-US decisions are reported in \S\ref{sec:res:policy}.

\subsection{Closed-Loop Skill Adaptation}
\label{sec:training}

The protocol-guided actor improves through an after-action adaptation
cycle that updates the Adaptive-Lesson bank from simulated epidemic
rollouts, leaving Protocol Constraints and Surveillance Skill
structures untouched and inference cost unchanged. After every $K$ closed-loop rollout episodes
$\{\hat{\tau}^{(k)}\}_{k=1}^{K}$ generated by the world model, a
teacher module performs retrospective analysis over trajectory
outcomes. The teacher focuses on policy-relevant events including
missed objectives, delayed interventions, epidemic rebounds,
capacity-limit violations, and instability in intervention timing.
Each analysed episode is distilled into an Adaptive Lesson defined
in Section~\ref{sec:policy}. The lesson bank is updated through additive insertion with
supersession,
\begin{equation}
\mathcal T_3
\leftarrow
\left(
\mathcal T_3
\setminus
\mathcal S(\ell)
\right)
\cup
\{\ell\},
\label{eq:lesson_update}
\end{equation}
where $\mathcal S(\ell)$ denotes lessons retired due to contradiction,
obsolescence, or lower confidence under newly observed evidence.
Unlike parameter optimisation in conventional policy learning, this
update preserves accumulated behavioural memory while allowing
outdated guidance to be replaced.
Each lesson is indexed by the tag set
$\tau(\mathbf{X}^t,\mathbf{S})$ of the rollout from which it was
distilled. The adaptation loop operates at a slower time scale than policy
generation: the actor selects interventions at each rollout
step, while lesson extraction and bank updates occur periodically
every $K$ episodes. This separation keeps online policy inference
unchanged while allowing the decision process to improve through
accumulated epidemic experience. The selection criteria, confidence
weighting, supersession rule and retrieval ranking are specified in
Appendix~\ref{app:teacher}.
\begin{table*}[!t]
\centering
\footnotesize
\setlength{\tabcolsep}{5pt}
\resizebox{\textwidth}{!}{%
\begin{tabular}{lrrrrrr}
\toprule
& \multicolumn{2}{c}{\textit{COVID-19 SEIR}} & \multicolumn{2}{c}{\textit{Influenza SEIR}} & \multicolumn{2}{c}{\textit{COVID-19 Real US (Omicron)}} \\
\cmidrule(lr){2-3}\cmidrule(lr){4-5}\cmidrule(lr){6-7}
Method & \shortstack{In-Dist.\\Peak / MAE} & \shortstack{OOD\\Peak / MAE} & \shortstack{In-Dist.\\Peak / MAE} & \shortstack{OOD\\Peak / MAE} & \shortstack{In-Dist.\\Peak / MAE} & \shortstack{OOD\\Peak / MAE} \\
\midrule
Compartmental-GP                & 0.062 / 0.051 & 0.558 / 0.533 & 0.354 / 0.331 & 1.139 / 1.084 & 0.332 / 0.178 & 1.102 / 1.068 \\
MechBayes (naive)               & 0.730 / 0.700 & 0.860 / 0.831 & 4.630 / 4.560 & 1.750 / 1.681 & 2.550 / 2.480 & 0.980 / 0.930 \\
MechBayes + Policy              & 0.470 / 0.448 & 0.490 / 0.461 & 2.970 / 2.905 & 0.980 / 0.929 & 1.700 / 1.620 & 0.630 / 0.590 \\
epidemia Mode A (naive)         & 0.680 / 0.648 & 0.750 / 0.709 & 4.300 / 4.185 & 1.500 / 1.431 & 2.400 / 2.310 & 0.910 / 0.860 \\
epidemia Mode B (pretrained)    & 0.210 / 0.194 & 0.250 / 0.228 & 0.770 / 0.723 & 0.500 / 0.460 & 0.430 / 0.390 & 0.310 / 0.280 \\
PAN-cODE$^{\dagger}$            & 0.105 / 0.084 & 0.462 / 0.504 & 0.607 / 0.548 & 0.956 / 1.016 & 0.553 / 0.366 & 0.912 / 0.903 \\
PAN-cODE + Policy$^{\dagger}$   & 0.105 / 0.091 & 0.427 / 0.448 & 0.602 / 0.592 & 0.869 / 0.901 & 0.531 / 0.319 & 0.848 / 0.806 \\
EARTH                           & 0.067 / 0.065 & 0.226 / 0.251 & 0.383 / 0.423 & 0.466 / 0.507 & 0.356 / 0.239 & 0.468 / 0.271 \\
\textbf{EARTH + Policy}         & \textbf{0.069 / 0.066} & \textbf{0.224 / 0.224} & \textbf{0.397 / 0.432} & \textbf{0.451 / 0.452} & \textbf{0.361 / 0.225} & \textbf{0.451 / 0.224} \\
\rowcolor{epiworldblue}
\textbf{EpiWorld}               & \textbf{0.063 / 0.061} & \textbf{0.210 / 0.237} & \textbf{0.361 / 0.398} & \textbf{0.435 / 0.479} & \textbf{0.339 / 0.211} & \textbf{0.414 / 0.237} \\
\bottomrule
\end{tabular}}
\caption{World-model forecast performance (12-week
autoregressive rollout, log1p per-100k). Each cell reports
\textit{Peak-MAE / MAE}; lower is better in both.
$^{\dagger}$PAN-cODE forecasts hospitalisations and deaths only.}
\label{tab:wm-performance}
\end{table*}

\begin{table*}[!t]
\centering
\footnotesize
\setlength{\tabcolsep}{10pt}
\begin{tabular}{lrrr}
\toprule
& \textit{COVID-19 Real US} & \textit{COVID-19 SEIR} & \textit{Influenza SEIR} \\
& (Omicron) & & \\
\cmidrule(lr){2-2}\cmidrule(lr){3-3}\cmidrule(lr){4-4}
Method & Hosp.\ $\Delta$ (8w) $\downarrow$ & Hosp.\ $\Delta$ (8w) $\downarrow$ & Hosp.\ $\Delta$ (8w) $\downarrow$ \\
\midrule
EpiPolicy-RL        & $-22.2\%$ & $-8.2\%$  & $-12.4\%$ \\
ADIOS (GGGP)        & $-33.4\%$ & $-12.5\%$ & $-18.6\%$ \\
\textbf{EpidRLearn} & $\mathbf{-37.5\%}$ & $\mathbf{-13.9\%}$ & $\mathbf{-20.9\%}$ \\
\midrule
\textbf{EpiWorld} (Base LLM)              & $-46.0\%$ & $-17.1\%$ & $-25.6\%$ \\
\textbf{EpiWorld} (with Static Skills)    & $-48.0\%$ & $-17.8\%$ & $-26.7\%$ \\
\rowcolor{epiworldblue}
\textbf{EpiWorld} (with Skill Evolution)  & $\mathbf{-59.3\%}$ & $\mathbf{-22.0\%}$ & $\mathbf{-33.0\%}$ \\
\bottomrule
\end{tabular}
\caption{Policy optimisation performance over 8-week rollouts, reported
as cumulative hospitalisation reduction against the reference
historical policy (lower is better). For scale, the reference
historical policy yields $406$ cumulative hospitalisations per 100k
over the 8-week horizon against $1{,}544$ under no intervention.
Policy-alignment diagnostics are reported separately in
Appendix~\ref{app:policy-alignment}.}
\label{tab:policy-performance}
\vspace{-10pt}
\end{table*}

\section{Experiments}
\label{sec:experiments}
\subsection{Data}
\label{sec:data}

We evaluate EpiWorld on three datasets. \textbf{Real US COVID-19}:
state-level weekly surveillance for 51 US states (Dec 2020 -- Feb
2022) from HHS Protect (hospitalisations) and JHU CSSE (cases,
deaths), spanning the Alpha, Delta, and Omicron waves; interventions
follow the 13-dim OxCGRT NPI grid~\citep{hale2021} (full grid in
Appendix~\ref{app:action-space}). \textbf{COVID-19 SEIR}: mechanistic
per-state SEIR(A) generator with policy-conditioned transmission and
GeoDS mobility coupling~\citep{kang2020multiscale}; real US CUSP
stay-at-home data calibrates simulator policy windows; the action
space is a 5-tier intensity scalar over 50 US states and 183 days.
\textbf{Influenza SEIR}: SEIRV variant with seasonal forcing, waning
immunity, and a vaccination compartment; per-state parameters are
calibrated against CDC FluView weekly
ILI~\citep{chakraborty2018howNot} so trajectories match real US state
seasonality; policy is a 2-dim action (advisory intensity,
vaccination tier), distinct from COVID's 13-dim grid, testing
cross-pathogen generality.

\subsection{Baselines and Evaluation}
\label{sec:baselines}

We compare against two families of baselines.
\begin{description}[font=\bfseries,leftmargin=0pt,labelindent=0pt,nosep,labelsep=0.5em]
\item[World-model forecasting.] Three architectural families of
epidemic forecasters: Bayesian compartmental (Compartmental-GP
\citep{qian2020compartmentalgp}, MechBayes~\citep{gibson2020mechbayes},
and epidemia~\citep{flaxman2020nature}), latent ODE (PAN-cODE
\citep{shi2024pancode}), and neural CDE (EARTH~\citep{wan2025earth}).
Each is evaluated in both a policy-blind variant (closest to the
published model) and a policy-conditioned variant (\textit{+ Policy})
that ingests per-week policy intensity through the same control
interface as EpiWorld, ensuring like-for-like comparison. In
Table~\ref{tab:wm-performance}, \textit{+ Policy} denotes the
policy-conditioned variant in each family; epidemia's \textit{Mode A}
fits per replicate while \textit{Mode B} pre-fits a global policy
effect on the training corpus. Per-baseline architectural details and
adaptations to our corpora are in Appendix~\ref{app:baselines}.
\item[Policy optimisation.] Three intervention-planning families:
PPO/actor-critic over the EpiPolicy simulator
(EpiPolicy-RL~\citep{epipolicyrl2023}), grammar-guided genetic
programming (ADIOS~\citep{adios_gggp}), and DQN with prioritised
replay on a contact-based SEIR (EpidRLearn~\citep{bampa2022epidrlearn}).
All policy baselines are trained and rolled out against the same
world model EpiWorld's policy actor uses, on the same evaluation
harness. Codebooks, reward functions, and training hyperparameters
are in Appendix~\ref{app:baselines}.
\item[Evaluation.] Each dataset reports three observables per state
per week (COVID: hospitalisations, cases, deaths per 100k; Influenza:
weighted ILI, flu hospitalisations, flu mortality per 100k), with
hospitalisations as the primary policy-relevant target. Forecast
quality is reported as mean absolute error (MAE) and peak-MAE on the
12-week autoregressive rollout. Policy quality is measured by a
goal-aligned trajectory score: cumulative hospitalisation reduction
under the proposed policy relative to the reference historical
policy. A policy-alignment
diagnostic (activation- and exact-match rate of proposed actions
against historical actions) is reported in
Appendix~\ref{app:policy-alignment} as a plausibility check on the
proposed action distribution rather than as a quality metric. Three OOD splits are
evaluated (temporal, state-wise, intervention-OOD); see
Appendix~\ref{app:datasets} for the mechanism behind each.
\end{description}

\section{Results}
\begin{figure*}[!t]
\centering
\includegraphics[width=0.90\linewidth, trim={0 4cm 0 0}, clip]{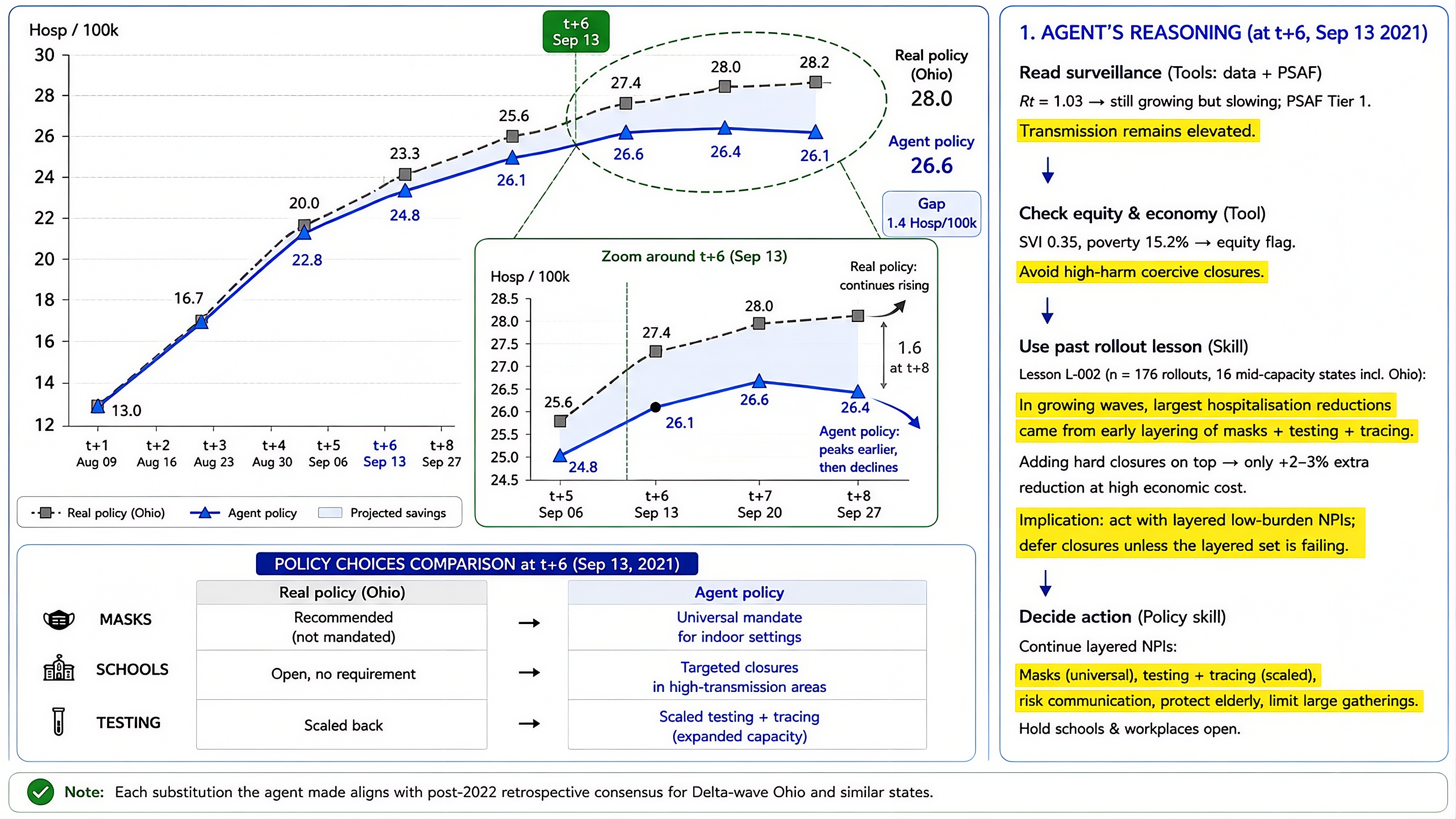}
\caption{Case study: Ohio during the Delta wave. Highlighted text indicates retrieved key reasoning steps.}
\label{fig:case-study-ohio}
\vspace{-13pt}
\end{figure*}
\label{sec:results}

\subsection{World Model Performance}
\label{sec:res:wm}

Table~\ref{tab:wm-performance} shows world-model performance.
EpiWorld outperforms all baselines on OOD Peak-MAE across the three
datasets, with gains of $\sim\!6\%$, $\sim\!4\%$, and $\sim\!8\%$ on
COVID-19 SEIR, Influenza SEIR, and Real US (Omicron) respectively
(versus the closest competitor, EARTH + Policy). The largest gain
comes on the real-data Omicron setting, where unseen-variant dynamics
expose the limits of baselines that rely on fixed mechanistic
structure; the smallest comes on Influenza SEIR, where the
simulator's explicit seasonal forcing and the simpler 2-dim action
space narrow the gap. The component ablation is repeated on influenza
in Appendix~\ref{app:flu-ablation}, where the same components remain
load-bearing. The Bayesian compartmental family is hampered
by rigid SEIR assumptions and improves only when policy is made
explicit (MechBayes + Policy roughly halves OOD MAE), while the
latent-ODE family lacks graph structure and trails uniformly. The
neural-CDE EARTH is the closest competitor through graph coupling
and reactive dynamics. EpiWorld outperforms both families by
combining graph coupling with policy-conditioned latent modulation
and delay-aware multi-channel decoding, giving stronger transfer to
unseen dynamics. On COVID-19 SEIR state-OOD the Peak-MAE advantage
over EARTH + Policy holds at paired Wilcoxon $p=0.039$ across $50$
held-out replicates.

A five-member ensemble of independently seeded world models supplies
a decision-time reliability signal: ensemble spread is strongly
rank-correlated with realised forecast error (Spearman $\rho=0.83$,
$p\approx0$), identifying unreliable forecasts before the error is
observable. The ensemble is under-dispersed --- nominal $95\%$
intervals cover $\sim\!68\%$ of outcomes overall and $73\%$ on the
hospitalisation channel --- so spread should be read as a relative
reliability signal rather than a calibrated interval.

\subsection{Policy Optimisation}
\label{sec:res:policy}

Table~\ref{tab:policy-performance} reports closed-loop policy
performance. EpiWorld with full Skill Evolution outperforms the RL
baselines on all three datasets, achieving a $\sim\!58\%$ further
reduction in cumulative hospitalisation relative to the strongest
baseline EpidRLearn ($-59.3\%$ vs.\ $-37.5\%$ on COVID-19 Real US).
The Base LLM already improves on every RL baseline before any
after-action signal, indicating that protocol-conditioned proposals
alone carry part of the gain.
Static skills add modest gains ($\sim\!1$--$2$ points); the Skill
Evolution cycle contributes the largest single component
($4$--$11$ points), peaking on the real-data Omicron setting. The
same pattern holds in the mean per-100k hospitalisation metric and on
the different flu NPI grid (Appendix~\ref{app:hosp-reduction}).

\begin{table}[t]
\centering
\footnotesize
\setlength{\tabcolsep}{4pt}
\begin{tabular}{@{}lrrr@{}}
\toprule
Method & In-model & Indep.\ sim & Ratio \\
\midrule
EpiPolicy-RL        & $-8.2$  & $-8.04$  & $0.98$ \\
ADIOS (GGGP)        & $-12.5$ & $-12.13$ & $0.97$ \\
EpidRLearn          & $-13.9$ & $-14.73$ & $1.06$ \\
\midrule
\textbf{EpiWorld} (Base LLM)      & $-17.1$ & $-18.30$ & $1.07$ \\
\textbf{EpiWorld} (Static Skills) & $-17.8$ & $-19.58$ & $1.10$ \\
\rowcolor{epiworldblue}
\textbf{EpiWorld} (Skill Evol.)   & $\mathbf{-22.0}$ & $\mathbf{-25.52}$ & $1.16$ \\
\bottomrule
\end{tabular}
\caption{Independent-simulator validation, COVID-19 SEIR state-OOD.
Cumulative hospitalisation reduction (\%) against the reference
historical policy, scored inside the learned world model $f_\theta$
and re-executed in the mechanistic simulator that generated the
corpus. Ratio is off-model over in-model. Real US is observational
data and admits no ground-truth generator.}
\label{tab:indep-sim}
\vspace{-8pt}
\end{table}

\begin{table}[t]
\centering
\footnotesize
\setlength{\tabcolsep}{3.5pt}
\begin{tabular}{lrrrr}
\toprule
Model & \shortstack{Base\\LLM} & \shortstack{w/ Static\\Skills} & \shortstack{w/ Skill\\Evolution} & $\Delta$\% \\
\midrule
\multicolumn{5}{l}{\textit{Proprietary}} \\
GPT-4o            & 0.1334 & 0.1371 & 0.0936 & $\mathbf{-29.8\%}$ \\
GPT-5-mini        & 0.1252 & 0.1468 & 0.1172 & $-6.4\%$ \\
GPT-5             & 0.1413 & 0.1468 & 0.1395 & $-1.3\%$ \\
\midrule
\multicolumn{5}{l}{\textit{Open-weights}} \\
Qwen2.5-7B        & 0.0884 & 0.0875 & 0.0692 & $-21.6\%$ \\
DeepSeek-R1       & 0.1332 & 0.1370 & 0.1174 & $-11.9\%$ \\
Llama-3.3-70B     & 0.1243 & 0.1247 & 0.0924 & $-25.7\%$ \\
\bottomrule
\end{tabular}
\caption{Cross-backbone ablation on COVID-19 SEIR (state-OOD). Cumulative hospitalisation per 100k under the
proposed policy over the 8-week post-intervention horizon (lower is
better). $\Delta\%$ is the relative reduction of the full Skill
Evolution configuration over Base LLM.}
\label{tab:backbone-ablation}
\vspace{-20pt}
\end{table}

Scoring every policy inside the world model it was optimised against
could reward exploiting the idiosyncrasies of $f_\theta$ rather than
genuine policy quality. We therefore re-executed all six methods in
the mechanistic SEIR simulator that generated the corpus, which
reproduces the released trajectories at maximum absolute error $0$.
Table~\ref{tab:indep-sim} reports both evaluations. The ordering is
preserved exactly across all six methods, each landing within
$\times 0.97$--$1.16$ of its in-model value, so world-model fidelity
does not alter the relative ranking. The margin over the strongest
baseline, EpidRLearn, is larger on independent ground truth
($+73\%$) than in-model ($+58\%$), and the ratio rises monotonically
with configuration quality: $f_\theta$ understates rather than
inflates the advantage of the closed-loop policy.

Because Tier-1 constraints bound the action set the actor reasons
over, their effect is measurable in the proposals themselves.
Across $2{,}640$ real-US decisions the envelope admits at most hand
hygiene in $42\%$ of decisions, mild measures in $21\%$ and a closure
in $37\%$, and never a full stay-at-home order. $59\%$ of proposed
actions fall within it; of the $41\%$ that do not, $39$ points are the
actor holding masking and testing in place while a wave declines and
the envelope has narrowed, at almost no burden. The remaining $2$
points are the only unsafe departures and are the decisions surfaced
for human approval.

Table~\ref{tab:backbone-ablation} isolates where the gain comes
from. The full skill cycle reduces hospitalisation across all six
backbones (average $\sim\!16\%$), while static skills alone leave
Base-LLM performance roughly unchanged: lessons are the load-bearing
component. GPT-4o achieves the largest reduction at $-30\%$; the
much smaller Qwen2.5-7B still reaches $-22\%$, indicating that the
gain is not limited by model scale only.

Replacing the actor with a fixed severity-proportional rule that
selects from the same skill library recovers only $59\%$ of the Base
LLM's reduction at matched intervention intensity
(Appendix~\ref{app:no-llm}): the gain comes from timing and targeting
rather than from stronger measures.

Adaptive Lessons generalise across states within an infection regime,
but are calibrated to the regime they are distilled from. Mined on
\{California, New York\} and applied to disjoint held-out OOD states on
the independent simulator, they reduce cumulative hospitalisation by a
further $9.0\%$ over the Base LLM (12/12 seeds), and by $4.3\%$ over
the tools-only configuration (11/12). Across regimes the sign
reverses: a bank distilled under Omicron, where the marginal return on
additional stringency is small, shrinks the expected effect size
($\delta=-0.234$); applied to a first wave in a fully susceptible
population ($R_0\approx1.7$--$2.9$) that shrinkage under-responds and
degrades outcomes by $12.1\%$ (Appendix~\ref{app:negative-transfer}).
We therefore mine a separate lesson bank per infection setting;
lessons are not universally transferable across arbitrary world
models. The same regime-dependence appears within a single setting:
on states whose baseline error is already low (FL, NJ, CA) the
multiplicative correction over-applies and per-state MAE regresses
(Appendix~\ref{app:per-state}), so the correction is most useful
where there is substantial error to remove. The
world-model design ablation is reported in
Appendix~\ref{app:additional}.

\subsection{Case Study: Ohio, Delta Wave}
\label{sec:res:case-study}

We analyse one held-out decision point to show how retrieved skills
inform policy. Under Ohio's historical interventions, the world
model reproduces the observed trajectory (MAE\,$=\,1.56$ per
100k). When the agent intervenes at $t{+}6$, the trajectory
bends downward, reducing peak hospital burden by $1.68$
per 100k ($\approx\!6\%$ at horizon). The proposed policy is not generated from hindsight labels: the
agent converges on layered mitigation (masking,
testing, tracing, targeted protection, and risk communication)
while avoiding broad closures. This broadly aligns with retrospective Delta-wave
analyses~\citep{brauner2021science,haug2020nhb,sachs2022lancet,talic2021bmj},
and the retrieved lesson L-002 prioritises layered low-burden NPIs
over stringency escalation, consistent with
public-health reviews reporting diminishing returns from blanket
closures~\citep{bjornskov2021cesifo}.
\vspace{-3pt}
\section{Conclusion}
\vspace{-3pt}
We presented EpiWorld, a closed-loop framework that grounds LLM policy agents in a learned epidemiological world model and protocol-guided skills to generate interpretable policies through counterfactual evaluation. Across all epidemic settings, EpiWorld improved out-of-distribution forecasting and closed-loop policy outcomes, suggesting that LLMs can serve as useful public-health decision-support tools under human oversight.
\section*{Limitations}
\label{sec:limitations}

EpiWorld is intended to assist public-health decision-makers with
learned policy proposals and counterfactual rollouts, not to automate
them; final decisions and accountability remain with human operators.
The evaluations reported here follow the standard world-model
protocol of forecast accuracy on out-of-distribution states and
counterfactual policy rollouts in the learned dynamics. Re-executing
every policy in an independent mechanistic simulator preserves the
relative ranking (\S\ref{sec:res:policy}), so the comparison does not
depend on the fidelity of $f_\theta$ alone; both evaluations remain
simulator-grounded and retrospective rather than prospective, and
cannot substitute for evaluation in a live deployment. In a real
setting the system would face behavioural fatigue, surveillance
heterogeneity, and the political and economic constraints that fall
outside the present goal space and the world model's training
support, and counterfactuals over action combinations absent from
historical data rely on extrapolation. Adaptive Lessons are
calibrated to the infection regime they are distilled from and do not
transfer across regimes, so a bank must be mined for each setting.
The five-member ensemble reported in \S\ref{sec:res:wm} ranks
forecast reliability well but is under-dispersed, so its intervals
should not be read as calibrated. Operational use would require
dedicated validation studies, equity audits, and continuous human
oversight.

\section*{Acknowledgments}
\label{sec:acknowledgments}

This work was supported in part by the National Science Foundation (NSF)
under Grant Nos.\ 2432418, 2414115, 2403312, 2318831, and 2521631; the
National Institutes of Health (NIH) under Grant Nos.\ R01AG089806 and
R01CA297856; and a Sony Research Award. The content is solely the
responsibility of the authors and does not necessarily represent the
official views of the NSF, NIH, or other funding organizations.

This work is supported in part by US National Science Foundation grants
\mbox{IIS-2312794} and \mbox{DBI-2412389}. Any opinions, findings, and conclusions or
recommendations expressed in this material are those of the author(s) and
do not necessarily reflect the views of the sponsor(s).

\bibliography{custom}

\appendix
\section{Datasets and Splits}
\label{app:datasets}

\paragraph{Real US COVID-19.} State-level weekly surveillance for 51
US states (50 + DC) over Dec 2020 -- Feb 2022, with interventions
from the 13-dimensional OxCGRT NPI grid~\citep{hale2021}.
Hospitalisations come from HHS Protect, cases and deaths from JHU
CSSE, mobility from GeoDS~\citep{kang2020multiscale}, and static
demographics inherited from PandemicLLM~\citep{du2024pandemicllm},
which assembles them from the American Community
Survey~\citep{us-census-acs}, the Commonwealth Fund state
health-system scorecard, and 2020 FEC election returns, together with
the ATSDR/CDC Social Vulnerability Index~\citep{atsdr-svi}.
The world model is trained on Dec 2020 -- Sep 2021 (Alpha and Delta
waves) and evaluated on the held-out Jan 2022 Omicron wave plus
eight subsequent weeks, giving a temporal-OOD setting.

\paragraph{COVID-19 SEIR (corpus \texttt{scenarios\_v2}).} A
per-state SEIR(A) generator with GeoDS mobility coupling, run under
policy windows derived from real US CUSP stay-at-home data. The
action grid is a five-tier intensity scalar
$\{0, 0.25, 0.5, 0.75, 1.0\}$. Training comprises 30 scenarios
$\times$ 5 replicates ($150$ trajectories) over 40 train states;
the in-distribution test set is 10 scenarios $\times$ 5 replicates
on the same train states; the state-OOD test set is 10 scenarios
$\times$ 5 replicates on 10 held-out states (DE, KS, MI, MN, MS,
NV, NJ, NC, SC, WA). Each replicate spans 183 days.

\paragraph{Influenza SEIR (corpus \texttt{flu\_scenarios\_v1}).} A
per-state SEIRV generator with seasonal $\beta(t)$ forcing, waning
immunity ($R\!\to\!S$, $V\!\to\!S$), and a vaccination compartment.
Per-state $(\beta_0, \alpha_{\text{season}}, \text{wILI scale})$ are
calibrated against CDC FluView weekly ILI for the 2017--18
season~\citep{chakraborty2018howNot}; FluView and FluVaxView data
parameterise the simulator and do not enter evaluation directly.
The action grid is two-dimensional (advisory intensity, vaccination
tier). Training and state-OOD test splits mirror COVID-19 SEIR (30
training scenarios, 10 OOD scenarios, 5 replicates each, 10
held-out states). Each replicate spans 240 days.

\paragraph{OOD splits.} \textbf{Temporal OOD} (Real US COVID-19):
training and test windows occupy different epidemic waves --
training uses Dec 2020 -- Sep 2021 (Alpha + Delta), while the test
window is the Jan 2022 Omicron wave plus eight subsequent weeks; the
Omicron-era dynamics (immune escape, faster generation interval,
milder severity) are absent from training. \textbf{State-wise OOD}
(COVID-19 SEIR and Influenza SEIR): a disjoint set of 10 states is
held out from training entirely; evaluation runs only on the
held-out states under the same disease and policy distributions as
training. \textbf{Intervention-OOD} (Influenza SEIR): test
trajectories are generated under policy configurations outside the
training distribution, with three sub-splits covering unseen
advisory intensities (\texttt{advisory OOD}), unseen vaccination
caps (\texttt{vaccination OOD}), and both combined
(\texttt{combined OOD}).

\paragraph{Legacy two-panel description.}
For reference, Table~\ref{tab:datasets-legacy} contrasts the COVID-19
surveillance panel with an Influenza-like Illness (ILI) dataset
preserved from an earlier two-panel framing. The ILI data was used
only as an observation-model sanity check and does not enter any
reported result.

\begin{table*}[!ht]
\centering
\footnotesize
\begin{tabularx}{\linewidth}{p{3.5cm}XX}
\toprule
& \textbf{COVID-19} & \textbf{Influenza-like Illness (ILI)} \\
\midrule
Used for &
forecast + policy &
observation model only \\
Period (ISO weeks) &
2020-W01 to 2022-W52 &
2012-W30 to 2013-W40 (PAHO), 2010-W40 to 2015-W20 (CDC) \\
Panels / records &
51 states $\times$ 156 weeks ($7{,}956$) &
15 country panels $\times \sim\!130$ weeks each \\
Cases / hosp / deaths &
JHU CSSE / HHS Protect / JHU CSSE &
ILINet wILI / n/a / n/a \\
Action grid &
13-dim OxCGRT~\citep{hale2021} &
none \\
Special structure &
per-week point estimates &
per-week revision history (initial $\to$ final) \\
Mobility, statics &
GeoDS~\citep{kang2020multiscale}, ACS, SVI~\citep{us-census-acs,atsdr-svi} &
\citet{chakraborty2018howNot} (PAHO, CDC); no static covariates \\
\bottomrule
\end{tabularx}
\caption{Legacy two-panel dataset description preserved for
reference. The ILI panel does not enter any reported result.}
\label{tab:datasets-legacy}
\end{table*}

\paragraph{Targets.} Each dataset reports three observables per
state per week, all $\log(1+x)$-transformed (metrics reported in the
same space). \emph{COVID-19}: hospitalisations, cases, and deaths
per 100k. \emph{Influenza}: weighted ILI \%, lab-confirmed flu
hospitalisations per 100k, and flu mortality per 100k.
Hospitalisations per 100k is the primary policy-relevant target
optimised by the closed-loop split. Closed-loop policy rollouts run
for 8 weeks per state.

\paragraph{Preprocessing.} Counts are converted to per-100k rates
using ACS state populations; negative weekly deltas are clipped to
zero; targets and mobility flows are $\log(1+x)$-transformed; static
indicators are z-scored on the train split. Missing weeks (median
$0.4\%$ on COVID-19; $1.7\%$ on Influenza) are linearly interpolated
within state with a missingness-indicator channel. Daily GeoDS flows
are aggregated to weekly OD counts and row-normalised per origin to
give the mobility matrix $\mathcal{G}$.

\section{Baselines: Adaptations and Hyperparameters}
\label{app:baselines}

\subsection{World-Model Forecasting Baselines}

\paragraph{Compartmental-GP~\citep{qian2020compartmentalgp}.} NumPyro
re-implementation of the published methodology. The published model
is a hierarchical two-layer Gaussian process with a compartmental
SEIR prior mean function, conditioned on per-country policy
indicators. With our 3-week warmup, the full hierarchical GP is
over-parameterised, so we approximate its temporal smoothness with an
AR(1) random walk on $\log R_t$ with mean reversion. Two variants:
\texttt{cgp\_naive} (AR(1) only, no policy term) and
\texttt{cgp\_policy} (AR(1) + global linear policy effect,
$\log R_t = \mathrm{AR}(1) + \beta \cdot \text{intensity}_t$).

\paragraph{MechBayes~\citep{gibson2020mechbayes}.} NumPyro
re-implementation of the published Bayesian SEIR model.
\texttt{mechbayes\_naive}: log-$\beta$ follows a per-state Gaussian
random walk during warmup, frozen post-warmup (closest to the
published non-parametric-$\beta$ formulation, policy-blind).
\texttt{mechbayes\_policy}: $\beta_t = \beta_0 (1 - 0.7 \cdot
\text{intensity}_t)$, matching the simulator's $\beta$-policy
mechanism.

\paragraph{epidemia~\citep{flaxman2020nature}.} NumPyro
re-implementation of the published semi-mechanistic Bayesian renewal
model. \texttt{epidemia\_modeA}: per-replicate fit. $\alpha$
(intercept) and $\beta$ (policy slope) sampled with priors and
inferred from the 3-week warmup of each test replicate.
\texttt{epidemia\_modeB}: hierarchical transfer.
$\alpha_\text{state}$ and $\beta_\text{global}$ pre-fit on the
training corpus via least-squares regression of weekly empirical
log-growth on policy intensity; at test time only initial conditions
and observation rates are sampled per replicate.

\paragraph{PAN-cODE~\citep{shi2024pancode}.} Latent ODE family:
GRU-ODE encoder $\to$ reparametrisation $\to$ Neural ODE $\to$
autoregressive decoder. Per-week policy intensity aliased into
PAN-cODE's stringency-index slot. Two variants: \texttt{uncond}
(encoder/ODE/decoder all policy-blind) and \texttt{+ Policy}
(per-week intensity concatenated to encoder input, latent $z_0$,
and decoder input at every step). Hyperparameters:
\texttt{latent\_dim=32}, \texttt{n\_units=50}, \texttt{n\_layer=3},
\texttt{lr=1e-3}, batch 128, ELBO with 3 MC samples, 80 epochs.

\paragraph{EARTH~\citep{wan2025earth}.} Neural CDE over interpolated
regional observations with multi-view spatial coupling. The published
model uses a static geographic adjacency and a DTW-derived semantic
adjacency to inject inter-region transmission pressure into the CDE
vector field. Lookback shortened from the published 8 weeks to our 3 weeks for
compatibility with the EpiWorld harness.
The policy-conditioned variant adds a 1-D policy encoder (inspired by
EpiWorld's PolicyEncoder) whose output modulates the CDE vector field
at each integration step.

\subsection{Policy-Optimisation Baselines}

\paragraph{EpiPolicy-RL~\citep{epipolicyrl2023}.} The published
method uses PPO and SAC over the EpiPolicy compartmental simulator
with a multi-objective reward combining health and intervention
cost. We re-implement the PPO actor against the EpiWorld world model
(the same dynamics our policy actor sees), using a Categorical actor
with a 5-tier intensity codebook for COVID-19 SEIR, the 13-dim
OxCGRT action for Real US COVID-19, and the 2-dim (advisory,
vaccination tier) action for Influenza SEIR.

\paragraph{ADIOS~\citep{adios_gggp}.} Grammar-guided genetic
programming for intervention discovery. The published method evolves
interventions as parse trees over a context-free grammar and
evaluates them inside the GEMS agent-based simulator. We swap the
simulator for the EpiWorld world model and adapt the grammar to each
dataset's action space (13-D OxCGRT, 5-tier intensity scalar, or
2-tier advisory$\times$vaccination). The fitness function combines
goal achievement with a stringency penalty. Population 50,
50 generations, tournament selection, 10\% mutation rate.

\paragraph{EpidRLearn~\citep{bampa2022epidrlearn}.} PPO (on-policy
actor--critic) over a contact-based, age-structured C-SEIR simulator
in the original paper, with a one-hidden-layer MLP ($128$ units,
$\tanh$, separate policy and value heads) and discount $\gamma{=}0.99$.
We re-implement the PPO actor against the EpiWorld world model for
apples-to-apples fairness, keeping the same policy/value
architecture and discount, and use the 5-tier intensity codebook
and reward function shared with EpiPolicy-RL.

\paragraph{Training protocol (all policy baselines).} 200K
environment steps against the world-model rollout harness, Adam at
$10^{-3}$, discount $\gamma=0.95$. Final policies are evaluated on
the held-out splits described in Section~\ref{sec:data} and
Appendix~\ref{app:datasets}.

\section{Feature Specification}
\label{app:features}

\paragraph{Static covariates.} 17 per-state indicators
($\mathbf{s}_i \in \mathbb{R}^{17}$), identical across all datasets
and z-scored on the train split (Table~\ref{tab:features-static}).

\begin{table}[!h]
\centering
\footnotesize
\setlength{\tabcolsep}{5pt}
\resizebox{\columnwidth}{!}{%
\begin{tabular}{ll}
\toprule
Feature & Description / source \\
\midrule
\texttt{Population}                   & ACS \\
\texttt{under\_20}                    & age-bracket fraction \\
\texttt{over\_65}                     & age-bracket fraction \\
\texttt{White}                        & race/ethnicity fraction \\
\texttt{Black}                        & race/ethnicity fraction \\
\texttt{medicaid\_coverage}           & fraction \\
\texttt{medicare\_coverage}           & fraction \\
\texttt{uninsured\_percent}           & ACS \\
\texttt{healthcare\_utilization}      & composite index \\
\texttt{poor\_health\_status}         & fraction \\
\texttt{adults\_at\_high\_risk}       & fraction \\
\texttt{poverty\_rate}                & ACS \\
\texttt{social\_vulnerability\_index} & CDC/ATSDR SVI \\
\texttt{healthcare\_access\_quality}  & composite index \\
\texttt{older\_at\_high\_risk}        & fraction \\
\texttt{dem\_percent}                 & political lean \\
\texttt{rep\_percent}                 & political lean \\
\bottomrule
\end{tabular}}
\caption{Static per-state covariates ($\mathbf{s}_i \in \mathbb{R}^{17}$).}
\label{tab:features-static}
\end{table}

\paragraph{Dynamic covariates.} 14 weekly channels per dataset
($\mathbf{X}_i^{\,t-T:t}\in\mathbb{R}^{14\times T}$),
$\log(1+x)$-transformed for the counts and clipped at zero
(Table~\ref{tab:features-dynamic}). Real US panels use observed
channels only; synthetic datasets additionally expose the
SEIR(V)-style latent compartments emitted by the simulator.

\begin{table}[!h]
\centering
\footnotesize
\setlength{\tabcolsep}{4pt}
\resizebox{\columnwidth}{!}{%
\begin{tabular}{llcc}
\toprule
Feature & Type & COVID & Flu \\
\midrule
\texttt{cases\_reported}             & per 100k         & \checkmark & \checkmark \\
\texttt{hosp\_reported}              & per 100k         & \checkmark & \checkmark \\
\texttt{death\_reported}             & per 100k         & \checkmark &            \\
\texttt{incidence}                   & per 100k         & \checkmark & \checkmark \\
\texttt{Rt}                          & Cori estimator    & \checkmark & \checkmark \\
\texttt{policy}                      & intensity scalar  & \checkmark & \checkmark \\
\texttt{vax\_curve}                  & cumulative cov.   &            & \checkmark \\
\texttt{previous\_infection\_12w}    & derived           & \checkmark & \checkmark \\
\texttt{Dose1\_Pop\_Pct}             & fraction          & \checkmark &            \\
\texttt{Series\_Complete\_Pop\_Pct}  & fraction          & \checkmark &            \\
\multicolumn{2}{l}{Latent $\{S,E,I,A,R\}$ (simulator-only, +5 ch.)} & \checkmark & \checkmark \\
\bottomrule
\end{tabular}}
\caption{Dynamic per-state covariates per dataset
($\mathbf{X}_i^{\,t-T:t}\in\mathbb{R}^{14\times T}$). Each dataset
sums to 14 channels: 9 observed + 5 latent.}
\label{tab:features-dynamic}
\end{table}

\section{Action-Space Specification}
\label{app:action-space}

The three datasets use different action interfaces, matched to what
each world model is conditioned on
(Table~\ref{tab:action-space}). Real US COVID-19 uses the published
13-D OxCGRT~\citep{hale2021} indicator panel for compatibility with
historical NPI records. COVID-19 SEIR collapses this to a 5-tier
intensity scalar, aligning the agent's decision with the 1-channel
stringency the simulator reacts to. Influenza SEIR splits a 2-D
continuous interface (advisory $\times$ vaccination cap) into four
tiers per channel. The OxCGRT panel groups into four envelopes used
by the Protocol-Constraint check (\Cref{sec:policy}): \textbf{personal
protective}: H1; \textbf{community measures}: C3M, C4M, H2, H3,
H6M; \textbf{closures and movement}: C1M, C2M, C6M, C7M, C8NV, C8V;
\textbf{vulnerable populations}: H8M.

\begin{table*}[!t]
\centering
\footnotesize
\setlength{\tabcolsep}{5pt}
\renewcommand{\arraystretch}{1.15}
\resizebox{\textwidth}{!}{%
\begin{tabular}{l l c p{0.55\textwidth}}
\toprule
Dataset & Dimension & Levels & Description \\
\midrule
\multicolumn{4}{l}{\textbf{Real US COVID-19}: 13-D OxCGRT indicator panel, ordinal NPI levels (increasing stringency)} \\
\midrule
& \texttt{C1M\_School closing}                 & $\{0,1,2,3\}$   & open / recommend / require some / require all \\
& \texttt{C2M\_Workplace closing}              & $\{0,1,2,3\}$   & open / recommend / require some / require all \\
& \texttt{C3M\_Cancel public events}           & $\{0,1,2\}$     & none / recommend / require \\
& \texttt{C4M\_Restrictions on gatherings}     & $\{0,1,2,3,4\}$ & no limit $\to$ $\le$10 persons \\
& \texttt{C6M\_Stay at home requirements}      & $\{0,1,2,3\}$   & none $\to$ minimal-exception lockdown \\
& \texttt{C7M\_Restrictions on internal movt.} & $\{0,1,2\}$     & none / recommend / restrict \\
& \texttt{C8NV\_Intl.\ travel (non-visitors)}  & $\{0,1,2,3,4\}$ & open $\to$ total border closure \\
& \texttt{C8V\_Intl.\ travel (visitors)}       & $\{0,1,2,3,4\}$ & open $\to$ total border closure \\
& \texttt{H1\_Public information campaigns}    & $\{0,1,2\}$     & none / coordinated public info \\
& \texttt{H2\_Testing policy}                  & $\{0,1,2,3\}$   & none $\to$ open-public testing \\
& \texttt{H3\_Contact tracing}                 & $\{0,1,2\}$     & none / limited / comprehensive \\
& \texttt{H6M\_Facial Coverings}               & $\{0,1,2,3,4\}$ & none $\to$ required in all public spaces \\
& \texttt{H8M\_Protection of elderly people}   & $\{0,1,2,3\}$   & none $\to$ extensive support and isolation \\
\midrule
\multicolumn{4}{l}{\textbf{COVID-19 SEIR}: 5-tier intensity scalar (aligned with WM's 1-channel stringency input)} \\
\midrule
& \texttt{intensity\_tier} & 5 tiers
 & \texttt{NO\_MEASURES} $\to$ \texttt{LIGHT} $\to$ \texttt{MODERATE} \newline $\to$ \texttt{STRICT} $\to$ \texttt{FULL\_LOCKDOWN} \\
\midrule
\multicolumn{4}{l}{\textbf{Influenza SEIR}: 2-D advisory + vaccination, four tiers per channel} \\
\midrule
& \texttt{advisory\_intensity} & 4 tiers
 & \texttt{NONE} ($0.05$), \texttt{LIGHT} ($0.15$), \texttt{MODERATE} ($0.30$), \texttt{STRICT} ($0.50$) \\
& \texttt{vax\_coverage\_cap}  & 4 tiers
 & \texttt{LOW} ($0.20$), \texttt{MID} ($0.35$), \texttt{HIGH} ($0.45$), \texttt{AGGRESSIVE} ($0.55$) \\
\bottomrule
\end{tabular}}
\caption{Action-space specification across the three datasets. The
agent selects a named tier; the world model receives the corresponding
scalar intensity. COVID-19 SEIR tiers map uniformly to
$\{0.00,0.25,0.50,0.75,1.00\}$; Influenza SEIR intensities are listed
in-line.}
\label{tab:action-space}
\end{table*}

\section{Skill Catalogue}
\label{app:skills}

Each tier's skills are listed in Tables~\ref{tab:t1-skills}--\ref{tab:t3-skills}.
Tier ordering mirrors the Method section (\Cref{sec:policy}).

\begin{table*}[!t]
\centering
\footnotesize
\setlength{\tabcolsep}{5pt}
\renewcommand{\arraystretch}{1.2}
\resizebox{\textwidth}{!}{%
\begin{tabular}{l p{0.45\textwidth} p{0.28\textwidth}}
\toprule
Skill & Function & Anchoring authority \\
\midrule
\texttt{tier1\_classify\_pandemic\_interval}
 & Classifies the six WHO pandemic intervals (Investigation, Recognition, Initiation, Acceleration, Deceleration, Preparation) from $(R_t, R_t\text{ CI}, \text{case trend}, \text{ICU}, \text{cluster detected})$; returns \texttt{max\_npi\_tier\_permitted}, which gates downstream NPI choice.
 & WHO pandemic-interval framework \\
\texttt{tier1\_vaccine\_allocation}
 & Enforces the priority ordering 1A $\to$ 1B $\to$ 1C $\to$ 2 $\to$ 3; flags \texttt{equity\_score}\,$<0.8$ for human review; prohibits price-, insurance-, or market-based allocation.
 & ACIP (US) / WHO SAGE \\
\texttt{tier1\_check\_npi\_proportionality}
 & Validates a proposed NPI bundle against the admissible tier envelope; requires legal authority for lockdown and movement restrictions; enforces a 14-day step-down minimum at $R_t<0.9$.
 & ECDC proportionality test \\
\texttt{tier1\_who\_outbreak\_investigation}
 & Enforces the WHO nine-step outbreak investigation ordering; blocks quantitative analysis before step~3 and control measures before step~6; mandates IHR notification at $\ge 5$ confirmed cases.
 & WHO IHR~\citep{who2005ihr} \\
\bottomrule
\end{tabular}}
\caption{Protocol Constraints ($\mathcal{T}_1$). Deterministic
predicates returning $\{0,1\}$ admissibility; computed once per
decision step.}
\label{tab:t1-skills}
\end{table*}

\begin{table*}[!t]
\centering
\footnotesize
\setlength{\tabcolsep}{5pt}
\renewcommand{\arraystretch}{1.2}
\resizebox{\textwidth}{!}{%
\begin{tabular}{l p{0.40\textwidth} p{0.30\textwidth}}
\toprule
Skill & Method (source) & Refinable parameters \\
\midrule
\texttt{tier2\_estimate\_rt}
 & Bayesian renewal-equation estimator for $R_t$~\citep{cori2013rt}.
 & serial-interval mean ($4.7$\,d), s.d.\ ($2.9$\,d), window $\tau$ (1\,w) \\
\texttt{tier2\_verity\_expected\_mortality}
 & Age-stratified IFR with case-ascertainment correction.
 & ascertainment ratio (0.3) \\
\texttt{tier2\_compute\_stringency\_index}
 & Nine-indicator OxCGRT normalisation to $[0,100]$~\citep{hale2021}.
 & --- \\
\texttt{tier2\_lookup\_svi}
 & ATSDR Social Vulnerability Index $\to$ quartile and equity directives~\citep{atsdr-svi}.
 & --- \\
\texttt{tier2\_compute\_growth\_rate}
 & Log-linear OLS for exponential growth rate $r$, doubling/halving time, and fit $R^2$.
 & stable band ($\pm 5\%$/week) \\
\texttt{tier2\_r\_from\_growth\_rate}
 & Wallinga--Lipsitch closed-form $R$ from growth rate $r$ and generation interval.
 & --- \\
\texttt{tier2\_classify\_psaf\_tier}
 & CDC Pandemic Severity Assessment Framework $\to$ severity tier 1--5.
 & hospitalisation severity thresholds, $R_t$ transmissibility bands \\
\bottomrule
\end{tabular}}
\caption{Surveillance Skills ($\mathcal{T}_2$). Analytical
estimators returning quantitative summaries; refinable parameters
may be adjusted by Adaptive Lessons within per-skill bounds.}
\label{tab:t2-skills}
\end{table*}

\begin{table*}[!t]
\centering
\footnotesize
\setlength{\tabcolsep}{5pt}
\renewcommand{\arraystretch}{1.2}
\resizebox{\textwidth}{!}{%
\begin{tabular}{l l p{0.55\textwidth}}
\toprule
Skill & Signature & Description \\
\midrule
\texttt{tier3\_retrieve\_lessons} & $(\tau, k)$
 & Returns the top-$k$ lessons ranked by tag overlap with the current context tags $\tau$, weighted by lesson confidence $\rho_\ell$. \\
\texttt{tier3\_apply\_lesson}     & $(\ell, \phi)$
 & Merges the parameter delta $\boldsymbol{\delta}_\ell$ into base parameters $\phi$, clamping to per-skill bounds $B_\kappa$; only keys in the refinable-parameter set are modulated. \\
\bottomrule
\end{tabular}}
\caption{Adaptive Lessons ($\mathcal{T}_3$). Retrieve-and-apply
operators over the lesson bank; structure is fixed, content
evolves through after-action analysis.}
\label{tab:t3-skills}
\end{table*}

The lesson schema, tag vocabulary, and a sample of the current lesson
bank appear in Appendix~\ref{app:lessons}.

\section{Key Agent Prompts}
\label{app:prompts}

\begin{figure*}[!t]
\begin{promptbox}[Protocol-Constraints system prompt (excerpt)]
\small
\textbf{Role.} You are an epidemiology policy agent operating under
the Protocol Constraints derived from WHO, CDC, and ECDC codified
guidelines. Protocol-Constraint outputs are non-overridable and
anchor every downstream decision.

\medskip
\textbf{Ordering rules (no exceptions).}
\begin{enumerate}\itemsep2pt\parsep0pt
\item \promptag{tier1\_classify\_pandemic\_interval} MUST be the first
  tool called in every session, before any other tool, and re-invoked
  whenever the epidemic state changes. Subsequent reasoning is gated
  by its outputs (\promptag{current\_interval},
  \promptag{max\_npi\_tier\_permitted}, \promptag{Rt\_status}).
\item \promptag{tier1\_who\_outbreak\_investigation} steps 1--3 must
  complete before any quantitative analysis.
\end{enumerate}
\end{promptbox}

\vspace{4pt}

\begin{promptbox}[Adaptive-Lessons after-action analyser prompt (excerpt)]
\small
\textbf{Role.} You are the Adaptive-Lessons analyser for an
epidemiology policy agent. Read a batch of rollout summaries plus the
current lessons bank, and propose new or refined lessons describing
cross-state behavioural patterns. Each lesson encodes a
\promptag{parameter\_delta} the agent will apply to a specific
Protocol-Constraint or Surveillance-Skill refinable parameter.

\medskip
\textbf{Hard rules.}
\begin{enumerate}\itemsep2pt\parsep0pt
\item Additive memory. You do not overwrite the bank. Same-sign
  refinements update an existing lesson in place; opposite-sign
  refinements require \promptag{retires: [L-XXX]} referencing the old
  lesson id, with a justification in \promptag{pattern\_observed}.
\end{enumerate}
\end{promptbox}

\vspace{4pt}

\begin{promptbox}[Reflection / prompt-evolution operator (excerpt)]
\small
\textbf{Role.} You are a prompt engineer optimising a system prompt
for an epidemic policy advisor. The advisor recommends a
13-dimensional NPI policy per region per week. The recommendations
are evaluated by a world model that predicts hospitalisation trends.
The goal is to minimise hospitalisations per 100k.

\medskip
\textbf{Task.}
\begin{enumerate}\itemsep2pt\parsep0pt
\item Diagnose specific weaknesses in the current system prompt that
  may have led to suboptimal recommendations.
\item Propose an improved system prompt addressing those weaknesses,
  under 250 words.
\end{enumerate}
\end{promptbox}
\end{figure*}

\section{Lesson Schema, Tag Vocabulary, and Examples}
\label{app:lessons}

\paragraph{Schema.} An Adaptive Lesson is a record with the
following fields:
\begin{itemize}\itemsep0pt
\item \texttt{id}: unique identifier (\eg{}\texttt{L-001}).
\item \texttt{applicable\_tags}: list of behavioural tags drawn from
  the vocabulary below.
\item \texttt{target\_skill}: identifier of the Protocol-Constraint
  or Surveillance Skill whose refinable parameter is adjusted.
\item \texttt{parameter\_delta}: dict of additive parameter shifts
  (clamped to per-skill bounds).
\item \texttt{regime}: the policy setting under which the cohort was
  observed.
\item \texttt{confidence} $\in[0,1]$: aggregated evidence strength
  across source states, computed as in
  Appendix~\ref{app:teacher}.
\item \texttt{sign\_agreement}: fraction of supporting observations
  agreeing in sign with the cohort median.
\item \texttt{source\_states}: states contributing observations
  ($\geq 2$ required).
\item \texttt{n\_observations}: number of contributing rollout
  observations.
\item \texttt{pattern\_observed}: free-text description of the
  divergence prompting the lesson.
\item \texttt{recommendation}: free-text statement of the action-level
  implication, used only for human inspection.
\item \texttt{retires}, \texttt{superseded\_by}: supersession
  metadata used to invalidate prior lessons whose sign conflicts with
  a newer one.
\end{itemize}

\paragraph{Tag vocabulary.} Tag sets
$\tau(\mathbf{X}^{t},\mathbf{S})$ are drawn from the following
namespaces:
\begin{figure*}[!t]
\begin{codebox}[Tag namespaces]
compliance:      low, mid, high
rebound:         fast, slow, none
capacity:        low, mid, high
vax-uptake:      slow, mid, fast
vax-ceiling:     low, mid, high
mandate-latency: slow, mid, fast
svi:             low, moderate_low, moderate_high, high
spread:          hub, rural, border, mixed
political:       dem, rep, mixed
psaf:            tier1, tier2, tier3, tier4, tier5
\end{codebox}
\end{figure*}

\paragraph{Example lessons.} Two regime-conditioned lessons from
the current bank (one COVID, one influenza), each combining a
cohort tag set with the active policy regime to produce an
action-level recommendation:

\begin{figure*}[!t]
\begin{codebox}[Example lessons (L-A: COVID, L-B: influenza)]
{
  "id": "L-A",                                        // COVID-19 SEIR
  "applicable_tags":
      ["access:high", "lean:dem", "pop:large", "svi:high"],
  "regime":  "stringency = MODERATE",
  "target_skill":  "tier2_classify_psaf_tier",
  "confidence":  0.46,
  "source_states":  ["California", "New York"],
  "n_observations":  21,
  "sign_agreement":  0.86,
  "pattern_observed":
    "SVI:high dense urban cohorts respond to MODERATE
     stringency as a lower-SVI cohort would respond to
     LIGHT -- transmission keeps growing.",
  "recommendation":
    "Same lever, one tier up: stringency MODERATE -> STRICT."
}

{
  "id": "L-B",                                        // Influenza SEIR
  "applicable_tags":
      ["access:high", "lean:dem", "pop:large", "svi:low"],
  "regime":  "advisory = ADV_MODERATE,  vax = VAX_HIGH",
  "target_skill":  "tier2_classify_psaf_tier",
  "confidence":  0.46,
  "source_states":  ["Illinois", "Pennsylvania"],
  "n_observations":  11,
  "sign_agreement":  1.00,
  "pattern_observed":
    "Vaccination is already saturated (VAX_HIGH) but a
     moderate advisory still lets transmission climb --
     the vaccination lever cannot substitute for contact
     reduction at this density.",
  "recommendation":
    "Different lever: hold vax = VAX_HIGH, escalate
     advisory ADV_MODERATE -> ADV_STRICT."
}
\end{codebox}
\end{figure*}

\section{Experimental Details}
\label{app:exp-details}

\paragraph{World-model instantiation.}
The world model in Section~\ref{sec:wm} is realised as a dual-encoder
Graph Transformer. The recency branch is a GRU over a 3-week window;
the delayed-effect branch is a Neural-CDE over the same 3-week window
with a causal cross-channel mixer (\textsc{DelayMix}). A learnable scalar
gate balances the two branches and their outputs are concatenated.
Policy actions enter through feature-wise (FiLM-style) modulation of
the latent state. Spatial coupling is a 4-layer adjacency-masked
Graph Transformer. A residual head predicts the additive update to
each observed channel (cases, hospitalisations, deaths per 100k
for COVID-19; cases and hospitalisations per 100k for influenza).
Training combines the latent-prediction and observation-reconstruction
losses defined in Equations~\ref{eq:wm-lat}--\ref{eq:wm-obs}.

\paragraph{World-model training.}
We train with AdamW (learning rate $3\!\times\!10^{-4}$, weight decay
$10^{-3}$, $10\%$ linear warmup, gradient clipping at $1.0$), batch
size $16$, and $5\,000$ steps. Targets are $\log(1+x)$-transformed
and standardised on the train split. Each model is trained on a
single NVIDIA A100 (Perlmutter). The best checkpoint by in-distribution
validation loss is used for all downstream evaluations.

\paragraph{Policy actor.}
The actor calls a chat-completion API for each backbone in
Table~\ref{tab:backbone-ablation} (GPT-4o, GPT-5-mini, GPT-5,
Qwen2.5-7B-Instruct, DeepSeek-R1, Llama-3.3-70B-Instruct) at
temperature $0.2$. Outputs are parsed from structured JSON with a
regex fallback to the \texttt{MODERATE} tier on parse failure. The
action grid is chosen per dataset as specified in
Table~\ref{tab:action-space}. Adaptive-Lesson retrieval injects the
top-$5$ lessons by tag overlap (Section~\ref{sec:policy}).

\paragraph{Closed-loop adaptation.}
Closed-loop rollouts run for an $8$-week horizon. Skill adaptation
runs after each rollout batch: the analyser proposes lessons from the
batch, and the bank is updated through additive insertion with
opposite-sign supersession (Equation~\ref{eq:lesson_update}). Protocol
Constraints and Surveillance Skill structures are not modified during
adaptation.

\paragraph{Evaluation.}
World-model accuracy is reported as Peak-MAE (absolute error at the
hospitalisation peak per state, per 100k) and mean trajectory MAE
on the OOD splits in Section~\ref{sec:data}. Policy performance is
reported as cumulative hospitalisation per 100k over an $8$-week
post-intervention horizon, averaged over the held-out states. All
policy methods are evaluated against the same world model and the
same initial conditions to isolate the effect of the actor.

\section{Additional Experiments}
\label{app:additional}

\begin{figure}[h]
\centering
\includegraphics[width=\columnwidth]{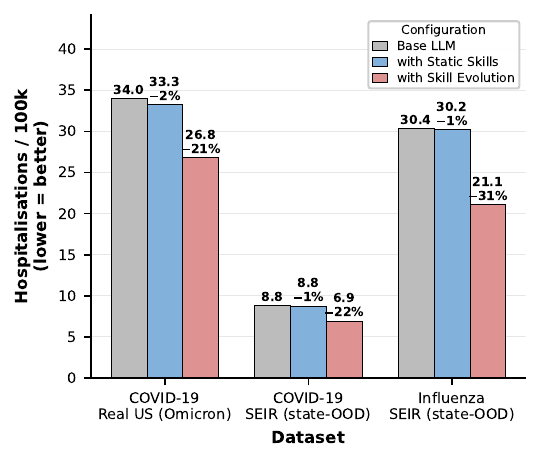}
\caption{Hospitalisation reduction (mean per-100k) across three
datasets and three skill configurations.}
\label{fig:reduction-app}
\end{figure}

\paragraph{Hospitalisation reduction (mean per-100k).}
\label{app:hosp-reduction}
Figure~\ref{fig:reduction-app} reports the \S\ref{sec:res:policy} ablation in the mean
per-100k hospitalisation metric. Static skills shift mean
hospitalisation by at most $2\%$ on any dataset, while the
after-action Skill Evolution cycle drives mean reduction to between
$21\%$ and $31\%$. The pattern is preserved on Influenza SEIR despite
its different NPI grid (CDC PMGR rather than OxCGRT), indicating the
gain is structural to the closed-loop skill cycle.

\paragraph{World-Model Design Ablation.}
\label{app:wm-ablation}
Table~\ref{tab:wm-ablation} reports a per-component ablation of the
EpiWorld world-model architecture on COVID-19 SEIR (Peak-MAE, log1p
per-100k). Dropping any single component degrades both
in-distribution and OOD Peak-MAE; the residual prediction head is
load-bearing, and replacing it with absolute next-state prediction
diverges. The largest single contributor to OOD generalisation is
the multi-stream SIR-coupled head; the dual encoder
(GRU + Neural-CDE) and the long lookback contribute next.

\begin{table}[!ht]
\centering
\footnotesize
\setlength{\tabcolsep}{3pt}
\begin{tabular}{lrr}
\toprule
Variant & \shortstack{In-Dist.\\Peak-MAE} & \shortstack{OOD\\Peak-MAE} \\
\midrule
\rowcolor{epiworldblue}
\textbf{EpiWorld (full)}             & \textbf{0.339} & \textbf{0.414} \\
$-$ dual encoder (CDE-only)          & 0.385 & 0.433 \\
$-$ dual encoder (GRU-only, 8w)      & 0.419 & 0.540 \\
$-$ long lookback (GRU 3w)           & 0.587 & 0.720 \\
$-$ DelayMixer (1-D causal mixer)    & 0.399 & 0.426 \\
$-$ multi-stream SIR-coupled head    & 0.554 & 0.734 \\
$-$ policy conditioning              & 0.421 & 0.585 \\
$-$ residual head (absolute)         & \multicolumn{2}{c}{diverges ($\geq\!8$)} \\
\bottomrule
\end{tabular}
\caption{World-model component ablation on COVID-19 SEIR (Peak-MAE,
log1p per-100k; lower better). Each row removes one component from
the full model.}
\label{tab:wm-ablation}
\end{table}

\paragraph{Influenza Component Ablation.}
\label{app:flu-ablation}
To test whether the architectural choices are specific to COVID-19,
we repeat the component ablation on the influenza corpus (two seeds
per configuration, influenza state-OOD). The dual temporal encoder is
load-bearing on influenza exactly as on COVID-19: removing the
Neural-CDE branch raises overall MAE from $0.487$ to $0.604$. The
DelayMixer gives the best influenza Peak-MAE ($0.441$ against
$0.463$ without it). The design is therefore not COVID-specific.

Scored head-to-head under an identical protocol, EpiWorld is lower
than EARTH + Policy on influenza overall rollout MAE ($0.541$ against
$0.581$) and on the hospitalisation channel ($0.491$ against
$0.525$), with peak error competitive. At the policy level, the
independent-simulator provenance gate passes (maximum absolute error
$3.4\times10^{-5}$) and Skill Evolution beats the Base LLM off-model
on the final-week metric ($-10.6\%$). The cumulative influenza margin
is smaller than on COVID-19, consistent with the simpler 2-dimensional
action space leaving less room for timing and targeting to matter.

\paragraph{Cross-Domain Negative Lesson Transfer.}
\label{app:negative-transfer}
Lessons mined under one world model do not necessarily port to a
different world model. We take the lesson bank
\texttt{lessons\_top20.json} mined under the real-US
\textsc{MechanismGraphWMReal} world model and apply it to
the synthetic \textsc{ScenarioGraphWMv15\_dual} on five held-out
OOD states. Pooled across states and replicates, the transferred
lessons \emph{degrade} performance by $+12.1\%$
(Table~\ref{tab:negative-transfer}).

\begin{table}[!h]
\centering
\footnotesize
\setlength{\tabcolsep}{5pt}
\begin{tabular}{lrrr}
\toprule
Held-out state & Base & + Real-US lessons & $\Delta$ \\
\midrule
KS &  12.51 &  12.64 & $+1.0\%$  \\
MN &   6.87 &   6.78 & $-1.4\%$  \\
NV &  57.74 &  74.79 & $+29.5\%$ \\
SC &  11.12 &  10.29 & $-7.5\%$  \\
WA &  88.24 &  93.28 & $+5.7\%$  \\
\midrule
\textbf{Pooled} & \textbf{35.30} & \textbf{39.56} & $\mathbf{+12.1\%}$ \\
\bottomrule
\end{tabular}
\caption{Cross-domain lesson transfer is negative. Lessons mined
under the real-US world model are applied to the synthetic SEIR
world model on five held-out OOD states. Mean hospitalisation per
$100$k; lower better.}
\label{tab:negative-transfer}
\end{table}

The mechanism is bias-direction-specific. The real-US WM
\emph{over-predicts} Omicron hospitalisations, so L-002's shrinkage
($\delta = -0.234$) corrects the bias and helps. The synthetic
v15\_dual WM \emph{under-predicts} on state-OOD, so the same
shrinkage \emph{amplifies} the under-prediction. Tier-3 lessons are
therefore not universally portable: they must be mined under the
target world model's own training distribution.

\paragraph{Per-State Breakdown.}
\label{app:per-state}
Table~\ref{tab:per-state-mae} reports per-state hospitalisation MAE
for the $20$ OOD-evaluated US states on the real-data Omicron split,
complementary to the aggregate hospitalisation metric in
Table~\ref{tab:policy-performance}. States are listed in decreasing
order of baseline MAE; bold marks the configuration with the lowest
MAE per state.

\begin{table}[!ht]
\centering
\footnotesize
\setlength{\tabcolsep}{4pt}
\begin{tabular}{lrrr}
\toprule
State & \shortstack{A\\(Base)} & \shortstack{D\\(+ Skills)} & \shortstack{F\\(+ Skill Evo.)} \\
\midrule
Arizona        & 23.34 & 25.03 & \textbf{14.24} \\
Indiana        & 20.01 & 18.84 & \textbf{10.86} \\
Wisconsin      & 19.16 & 17.06 & \textbf{10.44} \\
Pennsylvania   & 15.49 & 15.62 & \textbf{ 9.27} \\
Tennessee      & 15.24 & 16.18 & \textbf{10.03} \\
Michigan       & 14.07 & 14.63 & \textbf{ 6.28} \\
Ohio           & 13.25 & 13.14 & \textbf{ 8.66} \\
Massachusetts  & 12.66 & 12.32 & \textbf{ 9.19} \\
Missouri       & 11.75 & 11.77 & \textbf{ 8.67} \\
Georgia        & 11.27 & 10.48 & \textbf{ 9.03} \\
Texas          & 11.13 & 11.43 & \textbf{ 8.92} \\
North Carolina &  9.69 &  9.64 & \textbf{ 7.89} \\
New York       &  8.94 &  9.35 & \textbf{ 6.53} \\
Maryland       &  7.65 &  7.35 & \textbf{ 4.86} \\
Virginia       &  7.43 &  8.27 & \textbf{ 5.50} \\
Florida        &  7.08 & \textbf{ 6.38} &  9.32 \\
New Jersey     &  6.43 & \textbf{ 6.21} &  8.18 \\
California     &  5.70 & \textbf{ 4.42} &  8.44 \\
Illinois       &  5.45 & \textbf{ 5.20} &  7.41 \\
Washington     & \textbf{ 3.24} &  3.27 &  4.27 \\
\bottomrule
\end{tabular}
\caption{Per-state hospitalisation MAE (per 100k) on the real-US
Omicron OOD split. Skill Evolution improves $15$ of $20$ states and
gives its largest reductions on high-baseline-MAE Midwest states
(AZ, IN, WI). On the five lowest-baseline-MAE states (FL, NJ, CA,
IL, WA) the multiplicative correction over-applies and MAE
regresses; this regime-dependence is discussed in
Appendix~\ref{app:negative-transfer}.}
\label{tab:per-state-mae}
\end{table}

\section{Policy-Alignment Diagnostics}
\label{app:policy-alignment}

Policy alignment measures how far proposed interventions stay within
the distribution of actions taken in practice: the activation rate is
the fraction of decisions on which the proposed and historical action
agree on which measures are switched on, and the exact-match rate the
fraction on which they agree on the full action bundle. The score is a
plausibility diagnostic, not a measure of policy quality: the goal of
the framework is to improve on the reference historical policy, so
agreement with it is neither necessary nor sufficient for a good
outcome. Table~\ref{tab:policy-alignment} reports both rates for the
methods in Table~\ref{tab:policy-performance}.

\begin{table}[h]
\centering
\footnotesize
\setlength{\tabcolsep}{3pt}
\begin{tabular}{@{}lccc@{}}
\toprule
& \textit{Real US} & \textit{COVID} & \textit{Flu} \\
Method & Act./Ex. & Act./Ex. & Act./Ex. \\
\midrule
EpiPolicy-RL   & 43.1/22.3 & 92.3/10.4 & 38.1/17.2 \\
ADIOS (GGGP)   & 31.2/33.4 & 90.1/15.5 & 27.6/25.8 \\
EpidRLearn     & 31.9/33.7 & 90.4/15.8 & 28.2/26.0 \\
\midrule
EpiWorld (Base LLM)      & 66.1/50.3 & 97.2/23.3 & 58.5/38.8 \\
EpiWorld (Static Skills) & 69.2/53.8 & 97.7/25.0 & 61.2/41.5 \\
EpiWorld (Skill Evol.)   & 74.7/57.3 & 98.9/26.7 & 66.1/44.2 \\
\bottomrule
\end{tabular}
\caption{Policy-alignment diagnostics (activation / exact-match rate,
\%) for the methods in Table~\ref{tab:policy-performance}. Reported as
a plausibility check on the proposed action distribution, not as a
quality metric.}
\label{tab:policy-alignment}
\end{table}

On COVID-19 SEIR every method scores between $90.1\%$ and $98.9\%$
activation while cumulative hospitalisation reduction ranges from
$8.2\%$ to $22.0\%$, so alignment does not discriminate between
policies whose outcomes differ by a factor of $2.7$, and exact-match
never exceeds $60\%$ in any cell.

\section{No-LLM Rule-Based Selector}
\label{app:no-llm}

To separate the contribution of LLM reasoning from that of the
engineered skill library, we replace the policy actor with a fixed
severity-proportional rule that selects from the same library: the
Tier-2 $R_t$ and severity estimators set the intervention level, and
the Tier-1 constraints cap it exactly as they do for the LLM actor. No
language model is involved at any point in the decision.

\begin{table}[h]
\centering
\footnotesize
\setlength{\tabcolsep}{6pt}
\begin{tabular}{lrr}
\toprule
Selector & \shortstack{Cumulative\\reduction} & \shortstack{Mean\\intensity} \\
\midrule
Rule-based (no LLM) & $-21.4\%$ & $0.485$ \\
Base LLM            & $-36.5\%$ & $\approx 0.50$ \\
$+$ Skill Evolution & $-39.3\%$ & $\approx 0.50$ \\
\bottomrule
\end{tabular}
\caption{No-LLM ablation, COVID-19 SEIR state-OOD, evaluated on the
independent mechanistic simulator. All three selectors are scored on
an identical set of rollouts, so the values are directly comparable
within this table; they are computed on a matched subset and are not
on the same scale as Table~\ref{tab:indep-sim}.}
\label{tab:no-llm}
\end{table}

Removing the LLM roughly halves the reduction while the mean
intervention intensity is essentially unchanged ($0.485$ against
$\approx 0.50$), so the LLM does not obtain its advantage by
proposing more stringent measures. What it contributes is when and
where a comparable intervention budget is spent. The Base LLM alone,
without lessons or the after-action cycle, accounts for most of the
gap; Skill Evolution adds a further $2.8$ points.

\section{Teacher Module and Lesson Retrieval}
\label{app:teacher}

This section specifies the after-action analyser of
\S\ref{sec:training} and the retrieval operator of
Eq.~\ref{eq:t3-update}. The analyser learns which intervention levels
produce which outcomes for which kinds of state, and stores that as
adjustments to Tier-2 Surveillance-Skill parameters; Tier-1 Protocol
Constraints are never modified.

\paragraph{Selection.} Each closed-loop episode records how a state's
hospitalisation outcome responded to the stringency the actor
applied. Episodes are pooled into cohorts keyed by the state's
profile (population, social vulnerability, political lean, healthcare
access), the target skill, and the policy regime. A cohort yields a
lesson only if it draws on at least two distinct states, which
prevents single-state idiosyncrasies from entering the bank.

\paragraph{Confidence weighting.} Each cohort emits a delta
$\boldsymbol\delta_\ell$ on one refinable Tier-2 parameter, capturing
how that cohort's response differs from the skill's default. Its
confidence is
\begin{equation}
\rho_\ell
=
\underbrace{\left(1-\tfrac{1}{1+n/5}\right)}_{\text{evidence}}
\cdot
\underbrace{a_\ell}_{\text{sign agreement}}
\cdot
\underbrace{\left(1-\tfrac{1}{1+n_{\mathrm{states}}}\right)}_{\text{breadth}},
\label{eq:rho}
\end{equation}
where $n$ is the number of supporting observations, $a_\ell$ the
fraction of those observations agreeing in sign with the cohort
median, and $n_{\mathrm{states}}$ the number of contributing states.
All three factors lie in $[0,1]$, so $\rho_\ell\in[0,1]$.

\paragraph{Merge rule.} When a new observation batch refines an
existing lesson, confidence is blended by evidence weight,
\begin{equation}
\rho
\leftarrow
\frac{\rho_{\mathrm{old}} n_{\mathrm{old}}
      + \rho_{\mathrm{new}} n_{\mathrm{new}}}
     {n_{\mathrm{old}} + n_{\mathrm{new}}},
\label{eq:rho-merge}
\end{equation}
so $\rho_\ell$ is a running evidence-weighted mean rather than a
value overwritten by the most recent episode.

\paragraph{Contradictions.} Same-sign refinements merge in place via
Eq.~\ref{eq:rho-merge}. An opposite-sign refinement is rejected
unless it explicitly retires the prior lesson, which is then marked
superseded and retained rather than deleted. The bank therefore
cannot oscillate between opposite corrections, and the superseded
record preserves the audit trail.

\paragraph{Retrieval ranking.} At each decision the actor extracts
the state's context tags and ranks candidate lessons by
$n_{\mathrm{overlap}}\times\rho_\ell$, with tags unweighted. The
deltas of the top five lessons with at least two overlapping tags are
added to the Tier-2 parameters and clipped to $[-0.8, 0.8]$.

\paragraph{Convergence.} Every factor in Eq.~\ref{eq:rho} is bounded
and saturating in its argument, and Eq.~\ref{eq:rho-merge} is a
convex combination, so $\rho_\ell$ converges as observations
accumulate rather than drifting with episode order. Combined with the
$\geq 2$-state selection rule and the supersession policy, the bank
is stable under continued adaptation.

\end{document}